%% file: main.tex
\documentclass[conference]{IEEEtran}
\IEEEoverridecommandlockouts

\usepackage{cite}
\usepackage{amsmath,amssymb,amsfonts}
\usepackage{algorithmic}
\usepackage{graphicx}
\usepackage{textcomp}
\usepackage{subcaption}
\usepackage{enumitem}
\usepackage{float}
\usepackage{xcolor}
\PassOptionsToPackage{hyphens}{url}\usepackage{hyperref}

\def\BibTeX{{\rm B\kern-.05em{\sc i\kern-.025em b}\kern-.08em
    T\kern-.1667em\lower.7ex\hbox{E}\kern-.125emX}}

\usepackage{tikz}
\newcommand*\circled[1]{\tikz[baseline=(char.base)]{
            \node[shape=circle,fill,inner sep=0.5pt] (char) {\textcolor{white}{\small#1}};}}

\DeclareMathOperator*{\argmax}{arg\,max}

\usepackage{xspace}
\newcommand{\oursystem}{CoAdapt\xspace}

\begin{document}

\title{\oursystem: An LLM-based Framework for Adaptive Collaborative Perception in IIoT Robotic Swarms}

\author{
\IEEEauthorblockN{Houssam Hajj Hassan\IEEEauthorrefmark{1}, Antonia Maria Masucci\IEEEauthorrefmark{1}, Lynda Zitoune\IEEEauthorrefmark{2},  Salah Eddine Elayoubi\IEEEauthorrefmark{2}}
\IEEEauthorblockA{\IEEEauthorrefmark{1}Orange Innovation, Ch\^atillon, France.}
\IEEEauthorblockA{\IEEEauthorrefmark{2}	Université Paris-Saclay, CNRS, CentraleSupélec, Laboratoire des Signaux et Systèmes, France.\\
\emph{\{houssam.hajjhassan, antoniamaria.masucci\}@orange.com}\\
\emph{lynda.zitoune@l2s.centralesupelec.fr}, 
\emph{salaheddine.elayoubi@centralesupelec.fr}
}
}

\maketitle

\begin{abstract}
\input{abstract}

\end{abstract}


\section{Introduction}
\label{sec:intro}
\input{intro}

\section{Related Work}
\label{sec:related}
\input{related}

\section{The Adaptive Data Fusion Problem}
\label{sec:problem}
\input{problem}

\section{System Architecture}
\label{sec:architecture}
\input{solution}

\section{Experimental Evaluation}
\label{sec:evaluation}
\input{experiments}

\section{Conclusion and Future Work}
\label{sec:conclusion}
\input{conclusion}

\section*{Data Availability}
Our code, datasets and results are publicly available at: \url{https://github.com/houssamhh/coadapt}.

\section*{Acknowledgment}
This work is supported by the CANCUN project funded by the French National Research Agency (ref. ANR-24-CE25-3560).

\bibliographystyle{IEEEtran}
\bibliography{bibliography}
\end{document}

%% file: abstract.tex
Industrial IoT environments increasingly deploy autonomous mobile robots for tasks such as material handling, product assembly, or infrastructure inspection.
In such deployments, collaborative perception enables robots to share LiDAR observations and collectively construct a richer model of their environment than an individual agent could produce alone.
However, industrial environments are dynamic spaces where robot positions shift continuously, network bandwidth fluctuates, and the marginal contribution of robots to perception quality varies at runtime.
Existing collaborative perception approaches are designed for static participation assumptions and cannot adapt to these dynamics without sacrificing either detection precision or communication efficiency.
This paper presents \oursystem, an adaptive collaborative perception framework for IIoT robotic swarms in which a Large Language Model (LLM) serves as a runtime fusion controller, jointly deciding which robots participate in the fusion process and which fusion algorithm to apply based on the current spatial configuration and network state.
The LLM reasons over structured natural language descriptions of the scene derived from raw LiDAR point clouds, requiring no task-specific training and generalizing to unseen swarm topologies.
Evaluated on the OPV2V benchmark across 25 scenarios, our approach achieves a $38\%$ reduction in communication cost while maintaining detection precision comparable to static baseline approaches.

%% file: intro.tex
The proliferation of autonomous robotic systems in industrial environments has given rise to a new class of cyber-physical deployments, in which large numbers of mobile agents equipped with heterogeneous sensors, constrained computational resources, and wireless communication interfaces, must collectively perceive and act upon a shared environment~\cite{vermesan2022internet}. 
These robotic swarms are increasingly central to IIoT applications such as warehouse automation, infrastructure inspection, and autonomous logistics, where individual robots operate under severe perceptual limitations: restricted fields of view, sensor occlusions caused by obstacles or other agents, and range constraints that degrade single-agent detection at distance~\cite{debie2023swarm, wang2020v2vnet}.
Enabling such swarms to operate autonomously without continuous human supervision or centralized orchestration is a fundamental objective of modern IIoT system design, and a prerequisite for scalable deployment in dynamic, unstructured industrial environments.

Collaborative perception~\cite{chen2019fcooper, xu2022cobevt, xu2022opv2v} addresses these limitations by enabling robots to share sensory information, constructing a richer, more complete representation of the environment than an individual agent could produce alone. 
The dominant sensing modality in autonomous vehicle and robotic swarm deployments is LiDAR-based object-detection, which produces observations as a cluster of point coordinates. 
In this context, collaborative perception takes the form of cooperative data fusion, where point cloud observations from multiple robots are aggregated.
Three broad data fusion paradigms have emerged in the literature. Early fusion~\cite{chen2019fcooper} aggregates raw point clouds before any feature extraction, maximizing information sharing at the cost of high communication volume. Late fusion~\cite{xu2022opv2v} shares only final detection outputs (bounding boxes and confidence scores), minimizing bandwidth usage but discarding inter-agent complementarity. Intermediate fusion\cite{xu2022cobevt, liu2020who2com, liu2020when2com} shares learned feature representations at an intermediate network layer, achieving a favorable tradeoff between precision and communication cost.
Yet all three paradigms treat the fusion strategy as a static design choice, where the participating robot set and fusion algorithm are fixed at deployment, blind to runtime variations in network load or scene conditions.
This rigidity becomes particularly problematic in  dynamic IIoT swarm deployments, where environmental conditions (e.g., occlusions, robot positioning) vary at runtime, the network state fluctuates, and the marginal contribution of each robot to perception quality is neither fixed nor uniform.
Agents with overlapping fields of view introduce redundancy without improving coverage, whereas agents far from the region of interest add communication cost without detection benefit.
Moreover, such relationships vary widely at runtime: as robots move, the utility of each agent's contribution changes continuously, and a participation decision that was optimal may become suboptimal seconds later.
Addressing this challenge therefore requires jointly and continuously deciding which robots participate in the fusion process, and which fusion strategy to apply, in response to the current environment and network state.

Existing work addresses subsets of this problem in isolation, with the majority of them either fixing the fusion paradigm at design time~\cite{xu2022opv2v, xu2022cobevt} or adapting feature transmission within a fixed intermediate fusion architecture~\cite{hu2022where2comm, yang2023how2comm}, without considering runtime switching between fusion paradigms or joint adaptation of the participant set.
Static subset selection methods~\cite{golpayegani2018participant, liu2020when2com, liu2020who2com} determine participation at deployment time based on fixed topology assumptions, and degrade as robots move.
Network-aware fusion methods~\cite{hu2022where2comm} reduce communication volume but optimize independently of application-level precision requirements and runtime network fluctuations.
Learning approaches~\cite{li2021learning} have shown promise for adaptive cooperative perception but require extensive environment-specific training and generalize poorly to novel topologies.
However, none of these approaches jointly adapt participant selection and fusion strategy at runtime in response to changing network and environmental conditions.

In this paper, we propose \oursystem, an adaptive collaborative perception framework that jointly optimizes participant selection and fusion strategy at runtime, balancing detection precision against communication cost in dynamic IIoT swarm deployments.
Our approach introduces a Scene Abstraction Module that converts raw LiDAR point cloud data into structured natural language descriptions of the swarm's spatial configuration, enabling a Large Language Model (LLM) to serve as an autonomic fusion controller.
At each control cycle, the LLM receives a description of the current environment state, and produces a joint decision that includes the subset of robots that participate in the fusion process, and the fusion algorithm to apply.
\oursystem enables adaptive collaborative perception in IIoT systems through the following key contributions:
\begin{itemize}[leftmargin=*]
    \item [--] An adaptive collaborative perception framework that jointly adapts participant selection and fusion strategy at runtime in response to dynamic network and environmental conditions.
    \item [--] An LLM-driven fusion policy mechanism that reasons over structured natural language abstractions of LiDAR point cloud data, swarm geometry, and network state to produce joint participant selection and fusion algorithm decisions without task-specific training or exhaustive search.
    \item [--] An open-source prototype implementation evaluated on the OPV2V benchmark, demonstrating significant communication cost reduction while maintaining comparable detection precision.
\end{itemize}

The rest of the paper is organized as follows. Section~\ref{sec:related} provides a brief background about collaborative perception and surveys related work.
Section~\ref{sec:problem} presents the adaptive data fusion problem in dynamic robot swarms.
Section~\ref{sec:architecture} overviews the architecture of \oursystem, and Section~\ref{sec:evaluation} presents the experimental evaluation of our approach.
Section~\ref{sec:conclusion} concludes the paper with future directions.

%% file: related.tex
\begin{figure*}[t]
    \centering
    \begin{subfigure}[t]{0.32\textwidth}
        \centering
        \includegraphics[width=\linewidth]{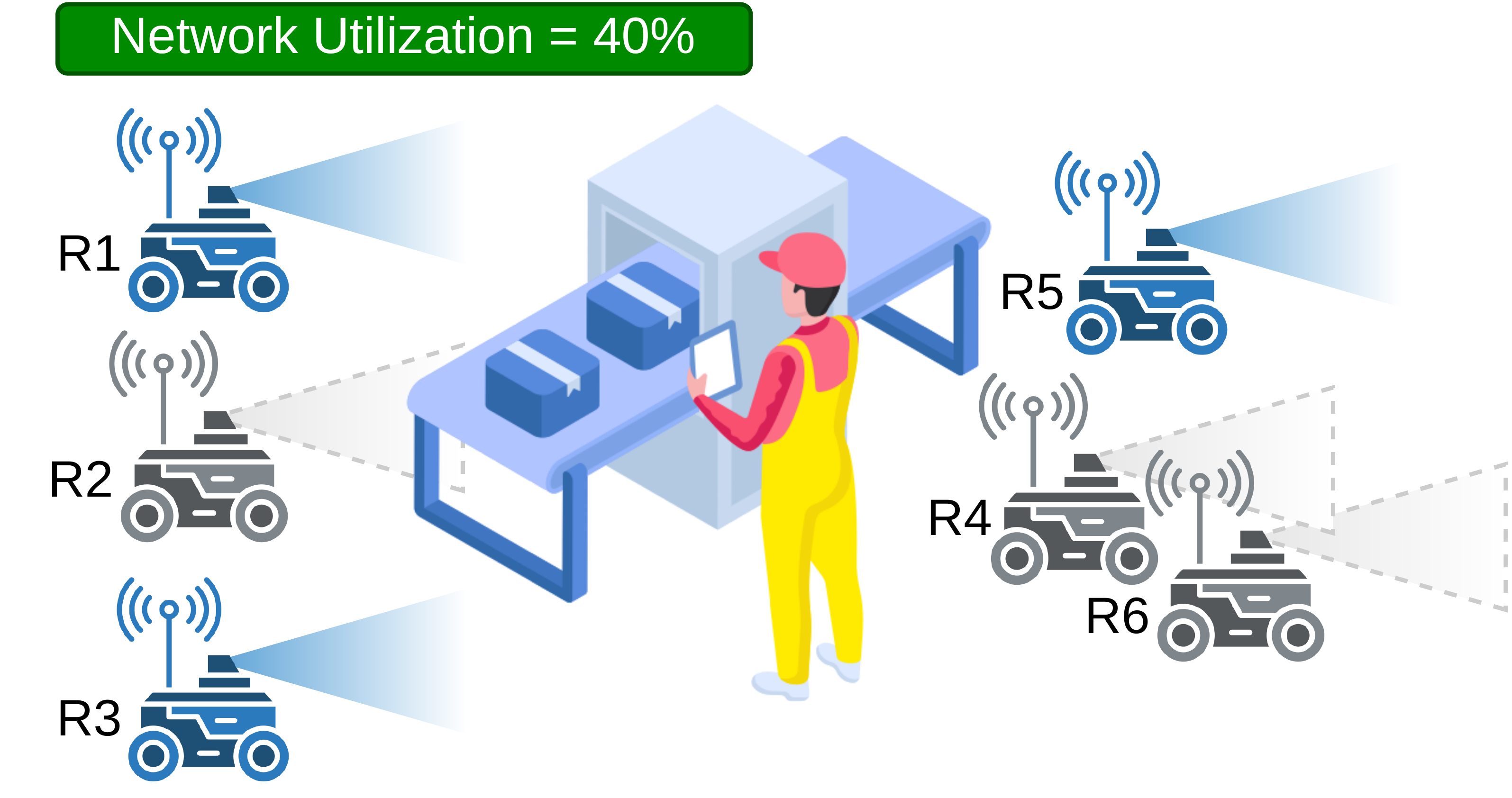}
        \caption{Geometric redundancy: $R5$ overlaps $R6$ and $R4$.}
        \label{fig:motivation-redundancy}
    \end{subfigure}
    \hfill
    \begin{subfigure}[t]{0.32\textwidth}
        \centering
        \includegraphics[width=\linewidth]{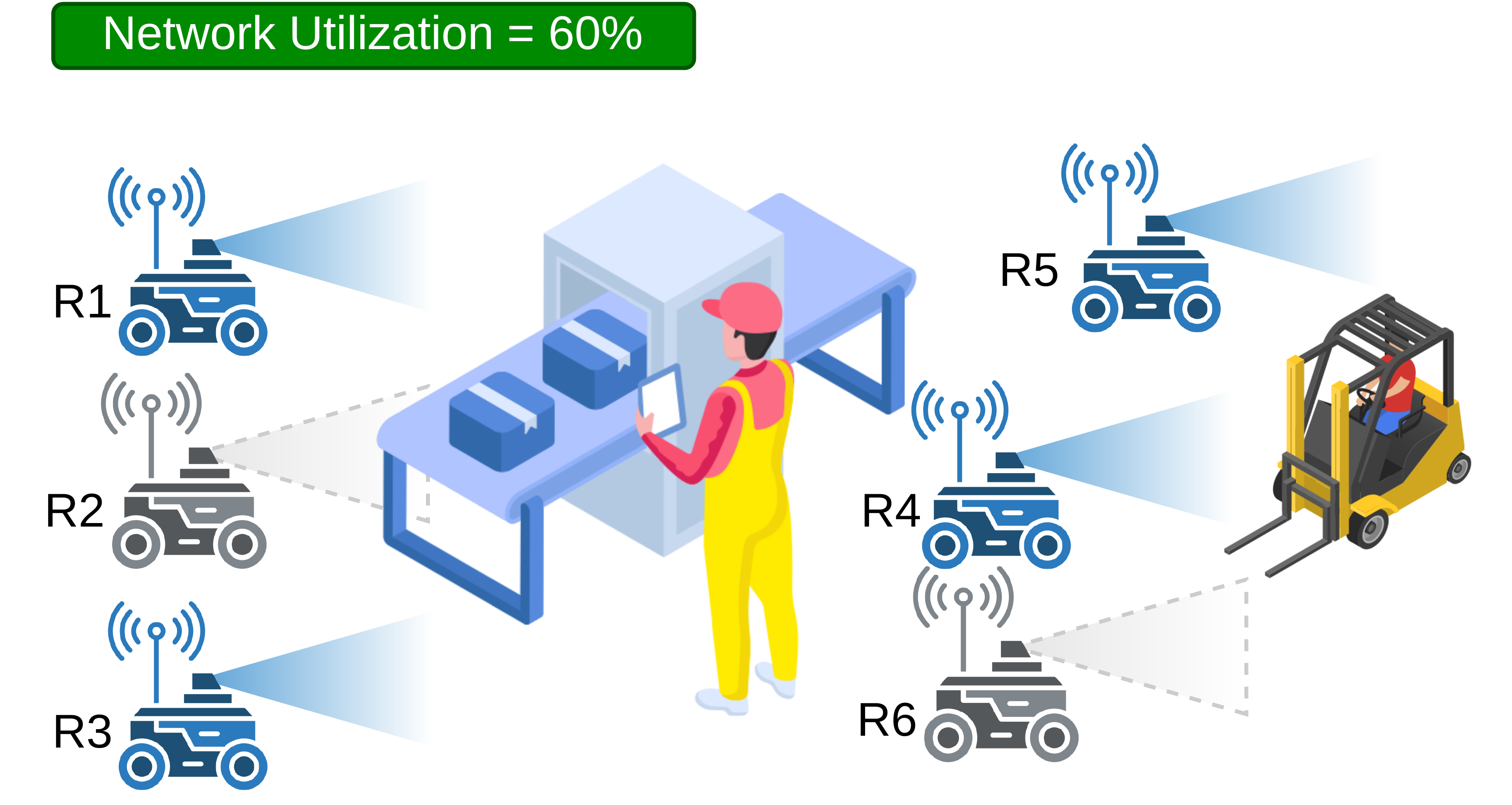}
        \caption{Dynamic occlusion: $R5$ uniquely covers a blind spot, making it indispensable.}
        \label{fig:motivation-occlusion}
    \end{subfigure}
    \hfill
    \begin{subfigure}[t]{0.32\textwidth}
        \centering
        \includegraphics[width=\linewidth]{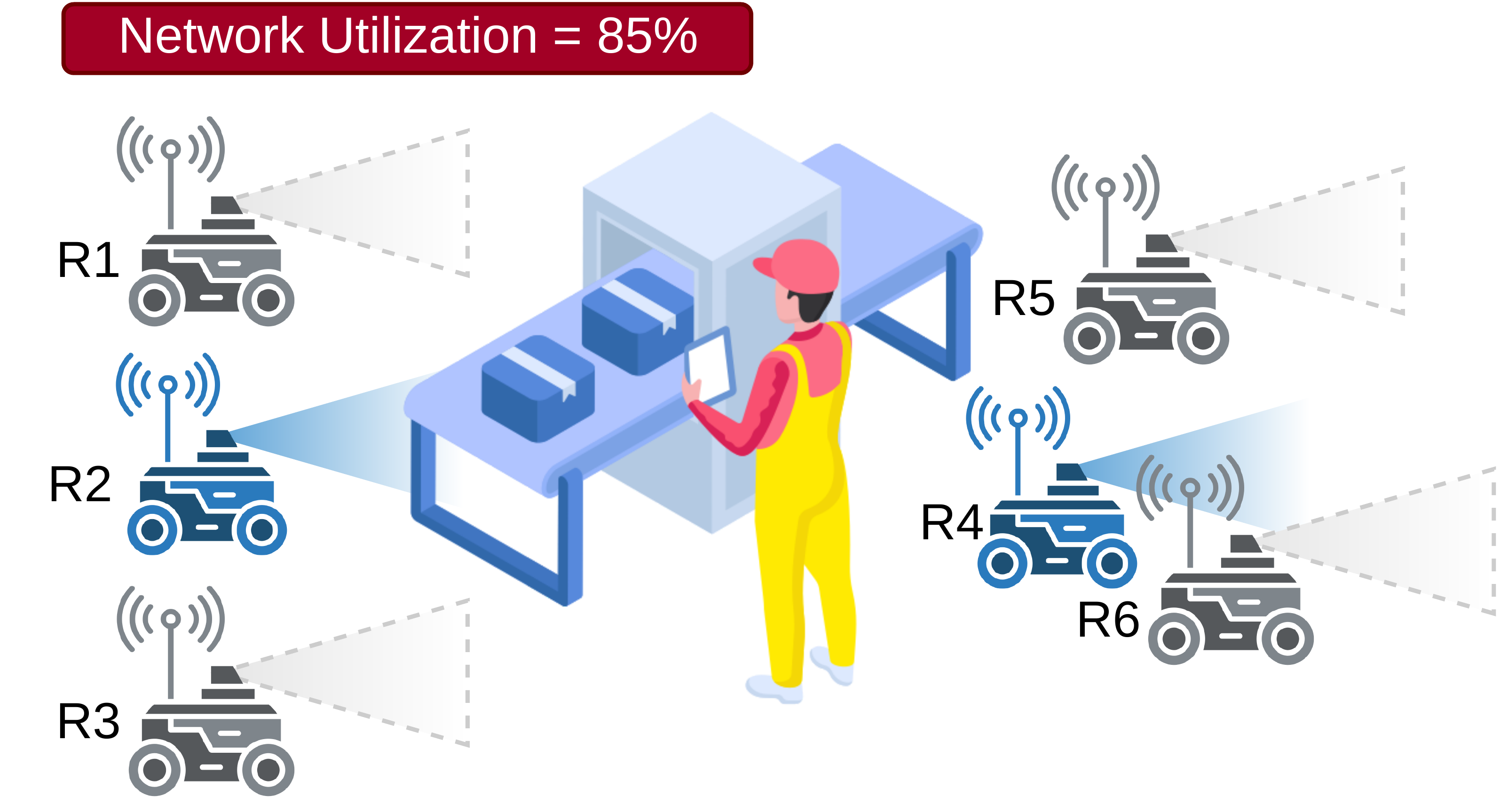}
        \caption{Network-QoS tension: bandwidth drop forces algorithm switch.}
        \label{fig:motivation-network}
    \end{subfigure}
    \caption{Three snapshots of an IIoT robot swarm exposing distinct challenges in adaptive collaborative perception. Blue robots are active participants; gray robots are excluded.}
    \label{fig:motivation}
\end{figure*}

This section provides first a brief background on collaborative perception, then surveys existing state-of-the-art approaches for communication-efficient collaborative perception, and self-adaptive robotic systems.

\paragraph{Collaborative perception and data fusion}

Collaborative perception has emerged as a promising paradigm for overcoming the inherent limitations of single-agent sensing in multi-robot systems~\cite{hadidi2018distributed, wang2020v2vnet, xu2022opv2v, han2023collaborative}.
By aggregating observations across agents, cooperative approaches significantly extend effective sensing range and mitigate occlusion effects that degrade single-robot perception. 
Three broad fusion paradigms have been established in the literature: early fusion, which aggregates raw sensor data (in the form of LiDAR point clouds, i.e., a set of data points in 3D space) before feature extraction~\cite{chen2019fcooper}; late fusion, which shares final detection outputs~\cite{xu2022opv2v}; and intermediate fusion, which exchanges learned feature representations at an intermediate network layer~\cite{xu2022cobevt, liu2020who2com, liu2020when2com}. 
Intermediate fusion has attracted significant attention as it achieves a favorable tradeoff between detection precision and communication overhead.
Recent work has begun to address the communication efficiency of collaborative perception more explicitly.
F-Cooper~\cite{chen2019fcooper} introduced the first feature-level fusion pipeline; V2VNet~\cite{wang2020v2vnet} applied spatially-aware graph neural networks for joint perception and prediction.
AttFuse~\cite{xu2022opv2v} and CoBEVT~\cite{xu2022cobevt} extend existing models with attention mechanisms, achieving state-of-the-art precision on benchmark datasets. 
Who2com~\cite{liu2020who2com} and When2com~\cite{liu2020when2com} learn to select communication partners and timing based on request-response matching. 
DiscoNet~\cite{li2021learning} employs knowledge distillation to compress intermediate representations, reducing transmission volume without sacrificing detection quality.

\paragraph{Communication-efficient collaborative perception}
A parallel line of work explicitly targets the communication cost of collaborative fusion.  Where2comm~\cite{hu2022where2comm} proposes spatial confidence maps that allow agents to transmit only perceptually critical features, achieving over $100,000\times$ reduction in communication volume on benchmark datasets while matching prior methods in detection precision. 
V2X-ViT~\cite{xu2022v2xvit} introduces heterogeneous multi-agent attention to reduce redundant transmissions.
SyncNet~\cite{lei2022latency} addresses latency-aware fusion under asynchronous communication by tackling temporal misalignment. 
CodeFilling~\cite{hu2024communication} addresses the communication bottleneck by using a codebook-based representation to transmit compact integer codes instead of high-dimensional feature maps across robots. 
Point Cluster~\cite{ding2025point} proposes a compact message unit that decouples low-level structural and high-level semantic information, enabling strategic point sampling to preserve scene structure at a lower communication cost than that of feature map transmission.
CoSDH~\cite{xu2025cosdh} combines supply-demand-aware region selection with an intermediate-late hybrid fusion mode, where late fusion compensates for accuracy degradation when bandwidth constraints make intermediate fusion insufficient.

While these approaches reduce communication overhead within a fixed fusion paradigm, none jointly adapts participant selection and fusion strategy at runtime in response to fluctuating network conditions and swarm dynamics.

\paragraph{Self-adaptive robotic systems}
Beyond perception-specific methods, a growing body of work addresses self-adaptation and autonomous decision-making in robotic systems more broadly.
ReBeT~\cite{alberts2024rebet} extends Behavior Trees with explicit quality requirement nodes to enable architecture-based self-adaptation in ROS2 robotic systems, allowing a robot to modify its software architecture at runtime to satisfy non-functional objectives such as energy consumption and safety.
Atik et al.~\cite{atik2024sustainability} focus on battery degradation-aware task and charge allocation algorithms that maximize mission revenue and battery lifespan in autonomous ground robots.
Swarmalators~\cite{schilcher2025swarmalators} addresses the practical challenge of implementing swarmalator dynamics on physical robots by deriving analytical bounds on the discrete-time step size that guarantee non-oscillatory convergence to emergent spatial patterns. 
Aguzzi et al.~\cite{aguzzi2025field} propose a decentralized runtime replanning approach grounded in aggregate programming, where continuously evolving fields represent mission state and enable robots to collectively adapt their task plan in response to failures without centralized coordination.
Large Language Models have also been used in robotics for task planning~\cite{huang2022language, kannan2024smart, ao2025llm}, scene understanding~\cite{liu2024vision}, and multi-agent coordination~\cite{yang2025autohma}. 
Notably, Nascimento et al.~\cite{nascimento2023selfadaptive} proposed integrating LLMs into MAPE-K-based multi-agent systems, demonstrating that GPT-based models can serve as autonomic controllers for self-adaptive behavior. 
In the context of autonomous driving, V2V-LLM~\cite{chiu2025v2v} explored multi-modal LLM reasoning for cooperative driving decisions. 
Li et al.~\cite{li2024genai} provide a comprehensive overview of generative AI within self-adaptive systems, specifically mapping LLM capabilities onto MAPE-K loop phases.

Despite this growing body of work, no existing approach applies LLM-based reasoning to the joint selection of fusion algorithm and participant subset in a collaborative perception system, nor does any prior work use LiDAR-derived point cloud data as the reasoning substrate for such decisions.

%% file: problem.tex
In this section, we first provide a motivating scenario to highlight self-adaptation  challenges in dynamic IIoT robot swarms.
We then formally present the adaptive data fusion problem.

\subsection{Motivating Scenario}

To highlight collaborative perception challenges in dynamic robotic swarms, we consider a smart manufacturing plant in the context of Industry 4.0 environments.
Consider an IIoT factory floor in which a swarm of several autonomous mobile robots (AMRs) $\{R_1, \ldots, R_6\}$ collaboratively perform object detection to support material handling operations.
Each robot is equipped with a LiDAR sensor and communicates over a shared wireless network.
The factory floor contains fixed infrastructure such as shelving units, conveyor belts, and machinery that create persistent occlusion zones, as well as moving obstacles such as forklifts and human workers that generate transient occlusions at runtime.

Figure~\ref{fig:motivation} illustrates three snapshots of this deployment, each exposing a distinct challenge that a static data fusion policy cannot resolve.
In Figure~\ref{fig:motivation-redundancy}, robots $R1$, $R2$, and $R3$
are tightly co-located near a conveyor belt, and therefore their fields of view substantially overlap.
Fusing all three observations from $R1$, $R2$, and $R3$ produces minimal coverage gain relative to fusing only $R1$ and $R3$, which occupy more complementary positions. 
The same applies for $R4$, $R5$, and $R6$, which are clustered next to each other.
Transmitting $R2$, $R4$, and $R6$'s full point clouds adds a significant load on the network without improving detection precision.
A static full-participation policy cannot identify this redundancy; an adaptive policy must reason over the current context and robot configuration to remove $R2$, $R4$, and $R6$ from the participant set.

Figure~\ref{fig:motivation-occlusion} shows how transient occlusions affect which robots participate in the fusion process.
As a forklift approaches $R4$'s field of view, $R4$'s contribution becomes uniquely valuable since no other robot is seeing the new obstacle.
At the same time, $R5$ navigates around the forklift and its field of view temporarily covers a blind spot invisible to all other robots.
In this context, excluding $R5$'s observation from the fusion process causes a missed detection. 
The optimal participant set at this timestep $\{R_1, R_3, R_4, R_5\}$ differs from that of Figure~\ref{fig:motivation-redundancy}.
In addition, as network utilization increases, sending compressed feature maps (intermediate fusion) instead of raw point clouds (early fusion) becomes the more appropriate approach.

Figure~\ref{fig:motivation-network} highlights a case where  telemetry traffic from industrial sensors on the factory floor leads to an increase in network utilization to $85\%$. 
Under these conditions, transmitting the raw point clouds or even compressed feature maps becomes infeasible within the real-time latency budget.
Late fusion reduces communication volume at the cost of slightly degrading detection precision.
A reduced participant set consisting of $\{R_2, R_4\}$ enables reducing communication data volume while satisfying detection precision constraints.
Identifying this combination requires reasoning jointly over the network state, the current robot positions, and the precision-communication tradeoff.
No existing collaborative perception system performs this joint reasoning at runtime.

These three snapshots define the core challenge that this paper solves, where the optimal fusion policy including which robots participate in the fusion process, is a function of the current environment state, the detection precision, and the network conditions, all of which change continuously during runtime operation.
Solving this challenge requires a system capable of autonomous and joint decision-making across all three dimensions.

\subsection{Problem Formulation}
We now formalize the collaborative perception problem for a robotic swarm operating under dynamic situations.
We consider a swarm of $N$ autonomous mobile robots $\mathcal{R} = \{r_1, r_2, \ldots, r_N\}$ operating on a shared factory floor environment.
Each robot $r_i$ is equipped with a LiDAR sensor and a wireless communication interface. At each discrete timestep $t \in \mathcal{T}$, robot $r_i$ acquires a point cloud observation $O_i(t) \in \mathbb{R}^{K_i(t) \times 3}$, where $K_i(t)$ is the number of points returned by the sensor at timestep $t$, and each point is represented by its 3D Cartesian coordinates $(x,y,z)$.
Robots communicate over a shared wireless network with capacity $C(t)$, which may vary over time due to competing traffic and channel conditions.

At time $t$, the full system is represented as  $\mathcal{S}(t) = (S^{env}(t), S^{net}(t), S^{app}(t))$, where:

\begin{itemize}
    \item \textit{Environment state} $S^{env}(t) = \{(x_i(t), \theta_i(t))\}^{N}_{i=1}$, where $x_i(t) \in \mathbb{R}^2$ and $\theta_i(t) \in [0, 2\pi)$ denote the position and heading of robot $r_i$. The environment state implicitly encodes inter-robot distances, field-of-view overlaps, and occlusion geometry induced by the swarm's current spatial configuration.
    \item \textit{Network state} $S^{net}(t) = C(t)$, representing the available communication bandwidth at timestep $t$.
    \item \textit{Application state} $S^{app}(t) = Q_{min}$, the minimum detection precision required by the application, expressed as average precision at a fixed Intersection over Union (IoU) threshold (AP@IoU). We treat this as a time-invariant requirement set at deployment.
\end{itemize}

Let $\mathcal{A} = \{\alpha_1, \alpha_2, \dots, \alpha_M\}$ denote the set of candidate fusion algorithms available to the system.
In our model, $\mathcal{A}$ includes early fusion, late fusion, and a set of intermediate fusion models.
A \textit{fusion policy} $\pi$ is a mapping:
\begin{equation}
    \pi : \mathcal{S}(t) \mapsto (P(t), \alpha(t))
\end{equation}

where $P(t) \subseteq \mathcal{R}, \left|P(t)\right| \geq 1$ is the selected subset of robots participating in the fusion process, and $\alpha(t) \in \mathcal{A}$ is the chosen fusion algorithm. 
The policy is re-evaluated at each control period and may produce a different decision as $\mathcal{S}(t)$ evolves.

To make the objective concrete, we define the network capacity $\mathcal{C}$ explicitly, which denotes the data volume (in bytes) transmitted over the network at timestep $t$.
For a participant subset $P(t)$ of size $n = |P(t)|$ under fusion algorithm $\alpha$:
\begin{equation}
    \mathcal{C}(P(t), \alpha(t)) = \sum_{r_i \in P(t)} \delta(\alpha, r_i(t))
\end{equation}
where $\delta(\alpha, r_i)$ is the per-robot transmission size under algorithm $\alpha$:
\begin{equation}
    \delta(\alpha, r_i) = \begin{cases}
        K_i \cdot 3 \cdot b_{float} & \textrm{if }\alpha = \textrm{early fusion} \\
        d_{feat} \cdot b_{float} & \textrm{if }\alpha \in \textrm{intermediate fusion} \\
        d_{det} \cdot b_{float} & \textrm{if }\alpha = \textrm{late fusion}
    \end{cases}
\end{equation}
where $b_{float}$ is the byte size of a floating-point value, $d_{feat}$ is the feature map size, and $d_{det}$ is the size of the detection output vector.

Let $\mathcal{Q} (P(t), \alpha(t))$ denote the object detection precision achieved by applying algorithm $\alpha$ to the fused observations of subset $P$ at time $t$, measured as AP@IoU. 
We formulate the objective as finding a policy $\pi^\ast$ that maximizes the expected detection precision while minimizing communication cost, subject to the application QoS constraints:
\begin{equation}
\begin{aligned}
    \pi^\ast & =  \argmax_{\pi} \mathbb{E} [ \mathcal{Q}(P(t), \alpha(t)) - \lambda \cdot \mathcal{C}(P(t), \alpha(t)) ] \\
    \textrm{s. t.} & \qquad \mathcal{Q}(P(t), \alpha(t)) \geq Q_{min} \quad \quad \quad \quad \forall t
\end{aligned}
\end{equation}
where $\lambda > 0$ is a scalar weighting factor that governs the precision-communication tradeoff.
Note that $\mathcal{C}$ depends on both the number and identity of participating robots $P(t)$, and the fusion algorithm chosen $\alpha(t)$, which determines the type and volume of data transmitted.

The decision space of $\pi$ at each timestep is $(2^N - 1) \times \left|\mathcal{A}\right|$, reflecting all non-empty subsets of $\mathcal{R}$ paired with each candidate fusion algorithm.
While manageable by exhaustive search in isolation, three properties make real-time optimization complex in practice.
First, the detection precision $\mathcal{Q}(P(t), \alpha(t))$ cannot be evaluated analytically.
It requires running the full fusion and detection pipeline, making exhaustive online evaluation infeasible within real-time latency budgets. 
Second, the relationship between $S^{env}(t)$ and $\mathcal{Q}$ is non-linear and environment-dependent: which robots are geometrically complementary depends on occlusion patterns that change continuously as robots move and the scene evolves.
Third, jointly reasoning over all three state components $(S^{env}, S^{net}, S^{app})$ simultaneously requires a policy capable of multi-objective reasoning, which neither heuristic selection rules nor single-objective optimization methods handle naturally.

These properties motivate the use of an LLM as the reasoning engine for $\pi$; rather than evaluating $\mathcal{Q}$ directly, the LLM approximates the mapping from $\mathcal{S}(t)$ to $(P(t), \alpha(t))$ through structured natural language reasoning over a compact scene abstraction, as described in Section~\ref{sec:architecture}.

%% file: solution.tex
\begin{figure}[t]
    \centering
    \includegraphics[width=\linewidth]{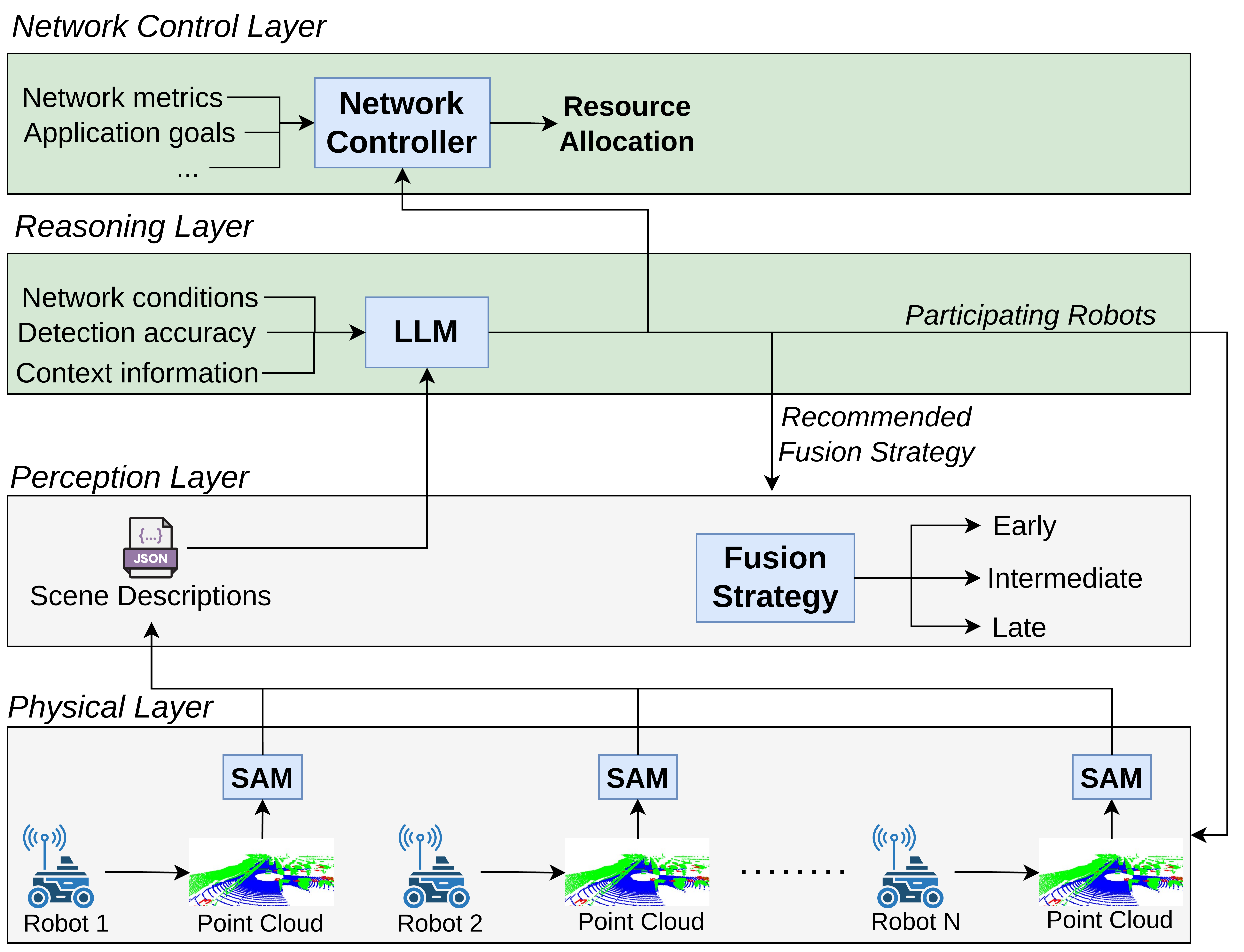}
    \caption{The \oursystem architecture.}
    \label{fig:architecture}
\end{figure}

\begin{figure}[htbp!]
    \centering
    \includegraphics[width=\linewidth]{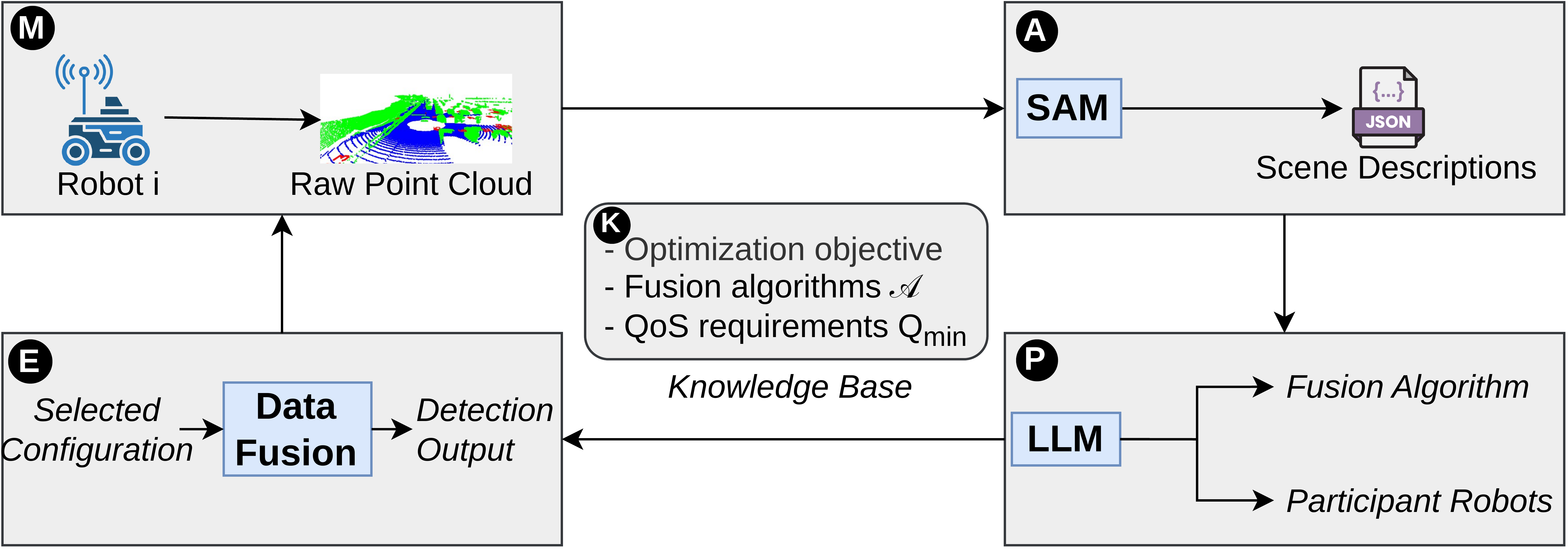}
    \caption{\oursystem's integration within the MAPE-K loop.}
    \label{fig:mapek}
\end{figure}

Existing collaborative perception systems treat fusion strategy and participant selection as fixed design-time decisions, leaving the system blind to the runtime dynamics that most affect performance, such as shifting swarm topology, transient occlusions, and fluctuating network conditions. We address this by framing collaborative fusion control as an autonomic decision problem and solving it with a large language model acting as the Plan component of a MAPE-K feedback loop~\cite{kephart2003vision}.
The key insight is that given a sufficiently structured description of the current system state, an LLM can reason jointly over geometric, application-level, and network objectives, producing a fusion policy $(P(t), \alpha(t))$ without exhaustive search and without retraining for new environments. 
This is made possible through our Scene Abstraction Module (SAM), which converts raw LiDAR point clouds and network telemetry into a compact JSON representation that exposes the information relevant to the fusion decision. 
Figure~\ref{fig:architecture} shows \oursystem's four-layer architecture. We describe each layer in detail below.

\paragraph{Physical Layer}
Each robot $r_i \in \mathcal{R}$ is equipped with a LiDAR sensor producing point cloud observation $O_i(t) \in \mathbb{R}^{K_i \times 3}
$ at regular intervals (e.g., 10Hz).
We consider that a leader robot (also called ego robot~\cite{hu2022where2comm, yang2023how2comm}) is the one executing the fusion policy, deciding which other robots to include as participants, and producing the final detection output.
Robots broadcast lightweight state beacons to the ego robot containing their current position $x_i(t)$ and heading $\theta_i(t)$.
In addition, we implement a \textbf{Scene Abstraction Module (SAM)} that enables each robot to produce structured descriptions of the scene it observes, based on the produced LiDAR observation.
The SAM addresses a fundamental interface problem where raw point clouds cannot be interpreted by LLMs. 
The SAM bridges this gap by transforming heterogeneous physical observations into a structured natural language representation that encodes the geometric and contextual information needed for fusion policy reasoning.
The SAM produces a description of objects perceived by robots including their position relative to the observing robot, their size and their density. 
More information about how the SAM is implemented is provided in Section~\ref{sec:evaluation}.
Robots periodically share their scene descriptions with the ego robot.
The ego robot also simultaneously monitors network conditions, measuring instantaneous available bandwidth $C(t)$ and per-link latency across the swarm. 
These raw observations, including physical observations and network metrics, are forwarded to the Perception Layer at each control cycle.

\paragraph{Perception Layer} 

The Perception Layer serves as the integration point between raw physical observations and the reasoning process.
Upon receiving the physical layer outputs at each control cycle, the ego robot fuses the individual scene descriptions from participating robots $P(t)$ with their reported positions $x_i(t)$ and heading $\theta_i(t)$, and the measured network metrics $C(t)$ into a unified scene representation.
This fusion produces a context-aware characterization of the environment that includes spatial relationships between robots, perceived object descriptions, and network conditions.
The resulting representation is structured as a natural language prompt and forwarded to the Reasoning Layer, where it serves as the informational basis for the LLM fusion policy.

\paragraph{Reasoning Layer}
The Reasoning Layer is responsible for adapting the data fusion algorithm and selecting participating robots at runtime.
A single LLM instance runs on the ego robot and serves as the autonomic fusion controller. 
At each control cycle, the ego robot constructs a prompt composed of three elements: (i)~a fixed \textit{system prompt} encoding the optimization objective in natural language (maximize the detection precision
subject to application goals and network constraints), along with the fusion algorithm vocabulary and output format specification; (ii)~the current scene descriptions from robots; and~(iii) a structured output constraint requiring the response to be valid JSON.
The LLM reasons jointly over swarm geometry, network state, application requirements, and prior performance, producing a fusion policy decision in the following format:
\begin{verbatim}
{
"participants": ["R1", "R3", "R4"],
"algorithm": "intermediate",
"reason": "R4 provides unique occlusion
coverage; bandwidth sufficient for
intermediate fusion with 3 agents"
}
\end{verbatim}

The \texttt{reason} field includes a human-readable explanation of the LLM's decision, providing a diagnostic signal for detecting system reasoning failures for IIoT operators.

\paragraph{Network Control Layer}

\oursystem includes a Network Control Layer operating above the Reasoning Layer, responsible for wireless resource management at a coarser timescale.
While the Reasoning Layer adapts the fusion policy to the current network state $C(t)$ by selecting participant subsets and fusion algorithms that fit within available bandwidth, it does so without modifying the network configuration itself.
The Network Control Layer closes this gap by actively adjusting network resources in response to application demands.
For instance, this can include assigning traffic priority to fusion data streams or reallocating bandwidth from non-critical channels when sustained congestion prevents the Reasoning Layer from meeting $Q_{min}$.
This layer operates at a slower timescale than the per-cycle decisions of the Reasoning Layer, intervening only when network conditions drift beyond what fusion policy adaptation alone can compensate for.
This two-layer design reflects a deliberate separation of concerns. 
The Reasoning Layer handles faster, environment-driven adaptation of the perception policy; the Network Control Layer handles slower, demand-driven adaptation of the network infrastructure.

\noindent \textbf{Integration within the MAPE-K loop.} \oursystem instantiates a MAPE-K control loop over the collaborative perception pipeline, as shown in Figure~\ref{fig:mapek}.
The Monitor (\circled{M}) phase collects raw observations from the physical layer at each control cycle that includes scene descriptions from participating robots, their positions, and measured network metrics. 
The Analyze (\circled{A}) phase constructs the unified scene representation in the Perception Layer, integrating perceptual and network state into a structured prompt.
The Plan (\circled{P}) phase is realized by the Reasoning Layer, where the LLM reasons over the current knowledge base to produce a fusion policy $\pi$. 
The Execute (\circled{E}) phase applies the selected fusion algorithm over the chosen participant subset, producing the final detection output. 
The shared Knowledge base (\circled{K}) includes the up-to-date optimization objective, set of available fusion algorithms $\mathcal{A}$, and QoS requirement $Q_{min}$, and is injected into the LLM system prompt at each control cycle.

%% file: experiments.tex
\begin{figure*}[t!]
    \centering
    \begin{subfigure}{0.48\textwidth}
        \includegraphics[width=\textwidth, trim=5 5 5 5, clip]{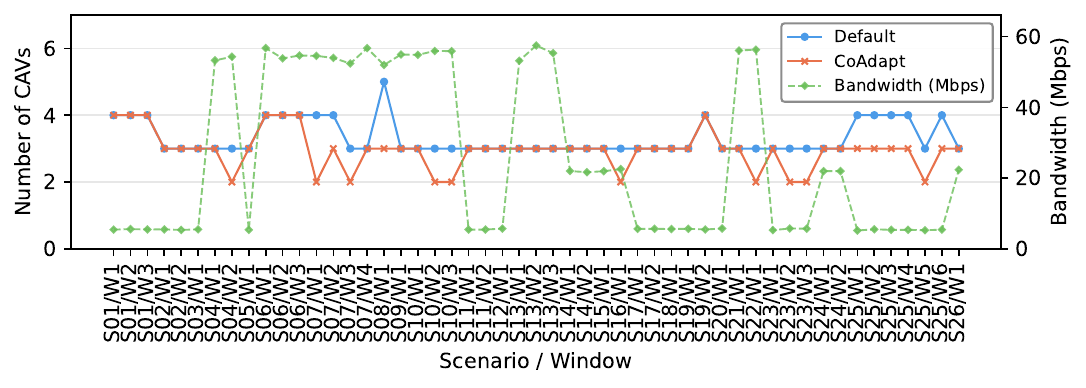}
        \caption{Gemma 4 31B.}
        \label{fig:cav_selection_gemma4}
    \end{subfigure}
    \hfill
    \begin{subfigure}{0.48\textwidth}
        \includegraphics[width=\textwidth, trim=5 5 5 5, clip]{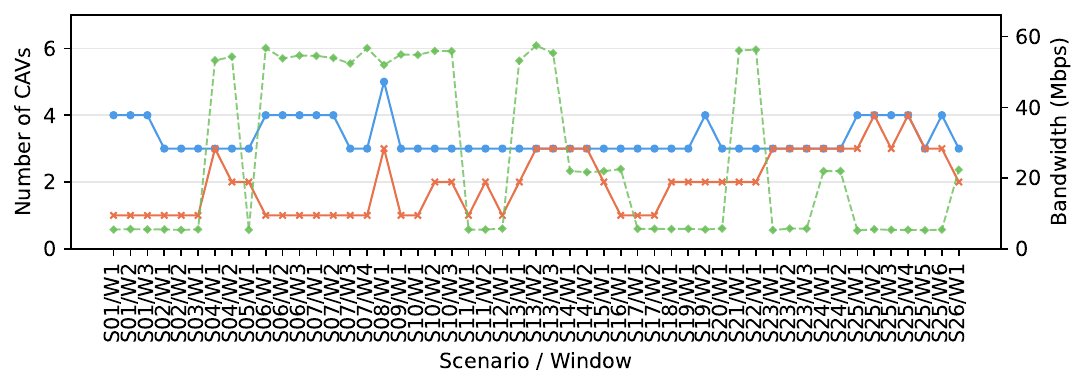}
        \caption{Llama3.3 70B.}
        \label{fig:cav_selection_llama}
    \end{subfigure}
    \vspace{0.5em}
    \begin{subfigure}{0.48\textwidth}
        \includegraphics[width=\textwidth, trim=5 5 5 5, clip]{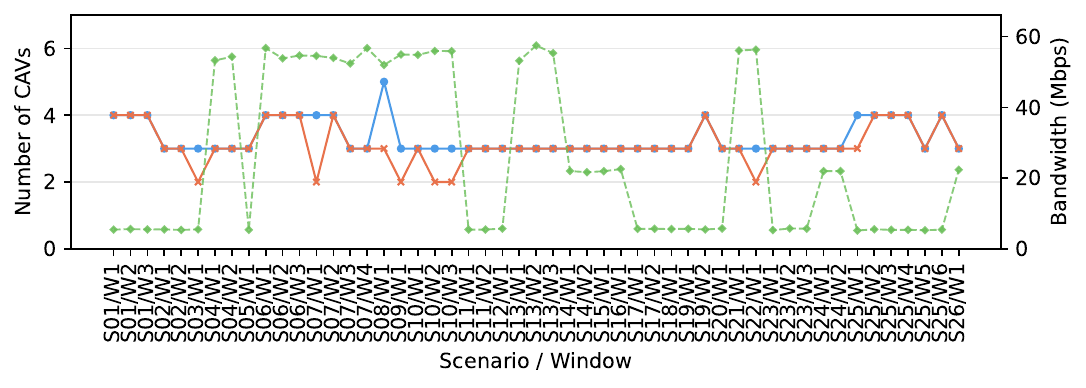}
        \caption{GPT-OSS 120B.}
        \label{fig:cav_selection_gpt120b}
    \end{subfigure}
    \hfill
    \begin{subfigure}{0.48\textwidth}
        \includegraphics[width=\textwidth, trim=5 5 5 5, clip]{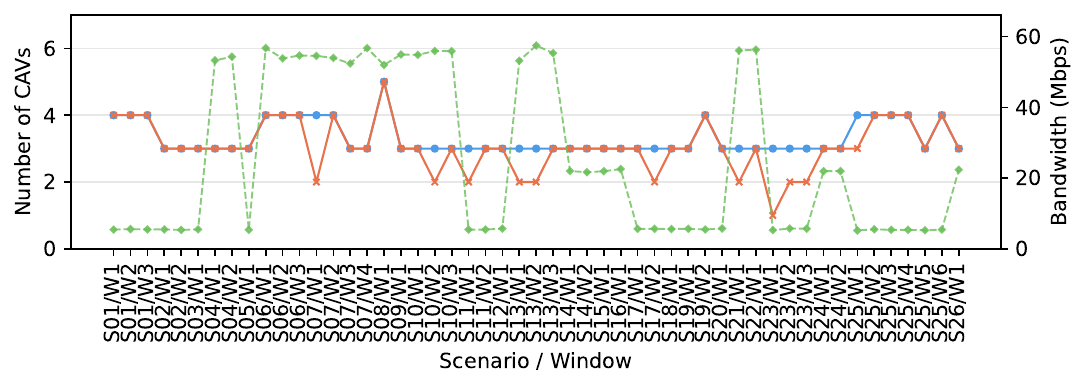}
        \caption{GPT-OSS 20B.}
        \label{fig:cav_selection_gpt20b}
    \end{subfigure}
    \caption{Number of selected CAVs vs. network bandwidth.}
    \label{fig:cav_selection}
\end{figure*}

This section presents the experimental evaluation of \oursystem.
We start first by describing the experimental setup including the dataset, data fusion models, implementation of the SAM, and LLM and bandwidth configurations.
We then evaluate \oursystem's ability to effectively select robots that participate in the fusion process under dynamic scene and network conditions.
We then proceed to show how \oursystem adapts the fusion strategy as a function of bandwidth, and finally compare the communication cost-precision tradeoff of \oursystem against rule-based and default approaches.
The code used for the experimental evaluation, along with the fusion models, datasets, configurations, and results, are publicly available at \url{https://github.com/houssamhh/coadapt}.

\subsection{Experimental Setup}

\noindent \textbf{Dataset.} We evaluate \oursystem on the OPV2V dataset~\cite{xu2022opv2v, opv2v}, a large-scale simulated collaborative perception benchmark dataset collected using the CARLA~\cite{carla} simulator across diverse urban and suburban scenarios. 
OPV2V is among the few publicly available datasets providing multi-vehicle LiDAR observations with up to six simultaneous agents per scenario.
This is a critical requirement for evaluating swarm-level fusion policy decisions, as datasets with a higher number of agent counts remain scarce in the collaborative perception literature~\cite{wang2025collaborative}.
The OPV2V dataset contains 73 scenarios simulated across 6 road types and 9 cities in CARLA.
Note that in our evaluation, we only consider scenarios containing 3 vehicles or more to effectively evaluate \oursystem's participant selection ability.

\noindent \textbf{Fusion Models.} We evaluate \oursystem's fusion policy against three static baselines representing each fusion paradigm, implemented within the OpenCOOD framework~\cite{opencood, xu2022opv2v}.
OpenCOOD provides a benchmark framework for evaluating collaborative perception algorithms against existing state-of-the-art approaches and datasets, including OPV2V.
In our evaluation, we select the Cooper model\cite{chen2019cooper} for early fusion, attentive fusion using the VoxelNet~\cite{zhou2018voxelnet} detection network for intermediate fusion, and late fusion using the Point Pillars model~\cite{lang2019pointpillars}.
These represent the best performing models (in terms of precision) available in OpenCOOD.
\oursystem selects among these three algorithms at runtime.
Note that \oursystem's architecture is agnostic to specific fusion models: any collaborative perception model can be substituted for the fusion backends evaluated here, without modification to the adaptation mechanism.
Fusion inference models run on an NVIDIA RTX 3090 GPU with 24GB of memory.

\noindent \textbf{Scene Abstraction Module (SAM).}  
The SAM is implemented using Open3D~\cite{zhou2018open3d} for point cloud processing. For each robot's LiDAR observation at each frame, the SAM first removes the ground plane via RANSAC-based plane segmentation~\cite{li2017improved}, isolating non-ground points. 
The remaining points are then clustered using DBSCAN~\cite{deng2020dbscan}.
Each remaining cluster is described by its 3D centroid position relative to the ego vehicle, its Euclidean distance to the ego vehicle, its axis-aligned bounding box dimensions, and its point density (points per unit volume). 
These per-object descriptions are serialized into structured natural language and appended to the LLM prompt alongside the robot's global position.

\noindent \textbf{LLM Configuration.} We evaluate 4 open-source LLMs: Gemma 4 (31B), GPT-OSS (20B and 120B), and Llama3.3 (70B).
All models are deployed locally on an NVIDIA H100 GPU with 80GB of memory. 
We select a control cycle window of 50 frames, which corresponds to a timestep of 5 seconds.
At each control cycle, the ego robot constructs a prompt comprising the current scene descriptions from all in-range vehicles, their positions, and the current network state, and queries the LLM for a fusion policy decision.

\noindent \textbf{Bandwidth Configuration.} Since OPV2V does not provide network measurements, we simulate time-varying bandwidth conditions using a two-component model designed to reflect realistic IIoT factory floor dynamics.
We map morning shift hours (06:00 am--12:00 pm) to a high-bandwidth tier of 55 Mbps, afternoon shift hours (12:00 pm--6:00 pm) to a medium tier of 20 Mbps, and evening shift hours (6:00 pm--12:00 am) to a low tier of 5 Mbps.
Time information is already encoded in scenario names in OPV2V and can readily be extracted.
To simulate network fluctuations, we add a sinusoidal fluctuation of amplitude 15\% of the base tier.

\subsection{Evaluating Participant Selection}

\begin{figure*}[thbp!]
    \centering
    \begin{subfigure}[t]{0.24\textwidth}
        \centering
        \includegraphics[width=\linewidth, trim=5 5 5 5, clip]{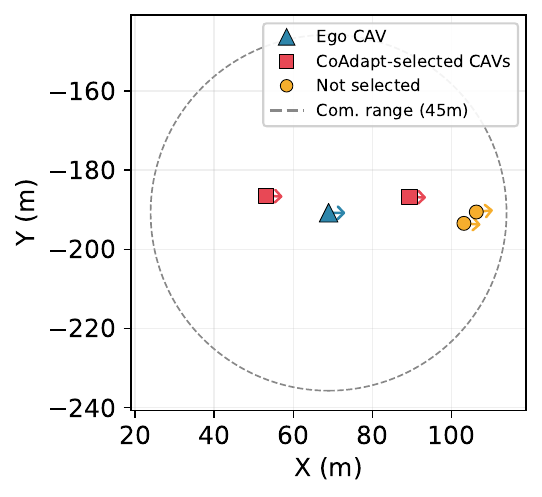}
        \caption{Gemma4.}
        \label{fig:cav_positions_gemma4}
    \end{subfigure}
    \hfill
    \begin{subfigure}[t]{0.24\textwidth}
        \centering
        \includegraphics[width=\linewidth, trim=5 5 5 5, clip]{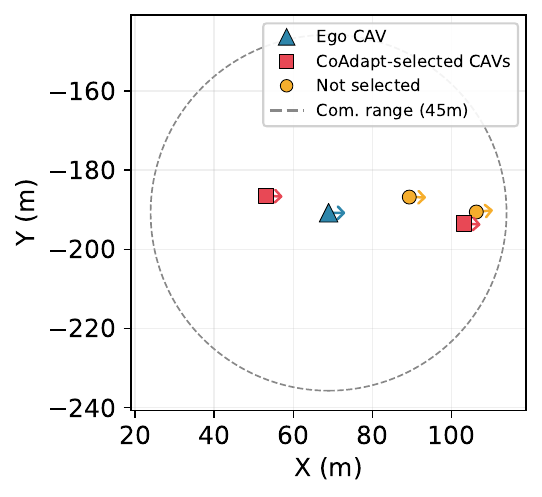}
        \caption{Llama 3.3 70B.}
        \label{fig:cav_positions_llama}
    \end{subfigure}
    \hfill
        \begin{subfigure}[t]{0.24\textwidth}
        \centering
        \includegraphics[width=\linewidth, trim=5 5 5 5, clip]{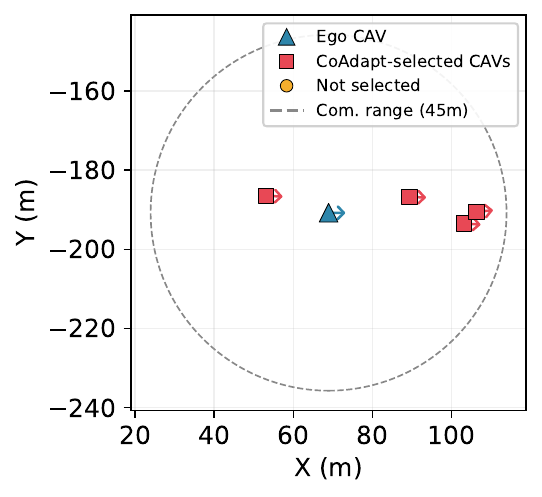}
        \caption{GPT-OSS 20B.}
        \label{fig:cav_positions_gpt20b}
    \end{subfigure}
    \hfill
    \begin{subfigure}[t]{0.24\textwidth}
        \centering
        \includegraphics[width=\linewidth, trim=5 5 5 5, clip]{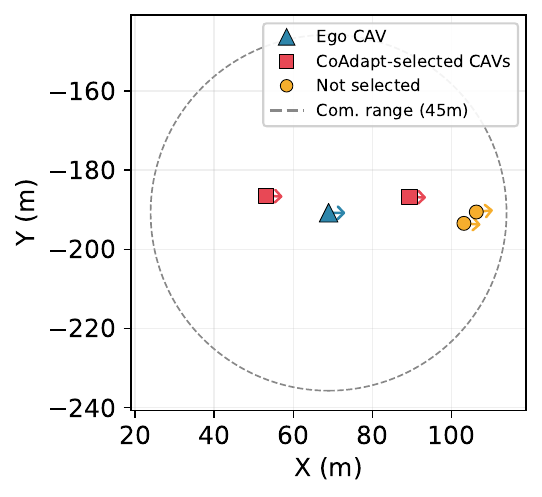}
        \caption{GPT-OSS 120B.}
        \label{fig:cav_positions_gpt120b}
    \end{subfigure}
    \caption{Spatial positions of \oursystem-selected CAVs.}
    \label{fig:cav_positions}
\end{figure*}

We first evaluate the ability of \oursystem to adapt participant selection in response to varying network conditions. Figure~\ref{fig:cav_selection} shows the number of Connected Autonomous Vehicles (CAVs) selected by each LLM across all evaluation scenarios, alongside the simulated available bandwidth.
The default baseline represents existing approaches that always include all in-range CAVs, irrespective of network conditions or scene geometry.

\oursystem consistently reduces participant count relative to the default full-participation baseline, pruning redundant CAVs based on their spatial contribution and the available network bandwidth.
All four tested LLMs consistently select fewer CAVs than the default approach across scenarios, achieving an average reduction of $26\%$ in participant count ($11\%$ for Gemma 4 31B, $7\%$ for GPT-OSS 20B, $6\%$ for GPT-OSS 120B, and $40\%$ for Llama3.3). 
This shows that the models respond to bandwidth variations, selecting fewer participants during low-bandwidth periods and maintaining higher participation when bandwidth is abundant. 
This network-aware behavior is most pronounced in Gemma 4 and Llama 3.3, which exhibit stronger correlation between bandwidth drops and participant reduction. GPT-OSS 20B and 120B show similar selection patterns to each other, suggesting that model scale alone does not determine selection behavior within the same model family.

\begin{figure*}[t!]
    \centering
    \begin{subfigure}{0.48\textwidth}
        \includegraphics[width=\textwidth, trim=5 5 5 5, clip]{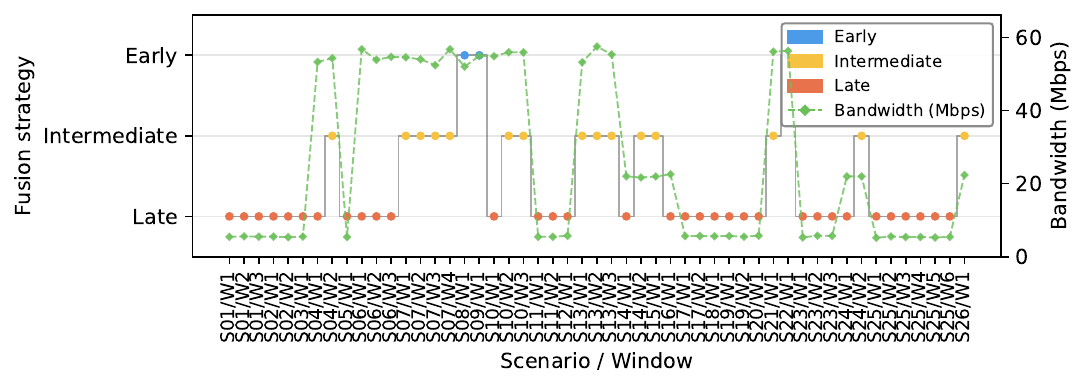}
        \caption{Gemma 4 31B.}
        \label{fig:fusion_selection_gemma4}
    \end{subfigure}
    \hfill
    \begin{subfigure}{0.48\textwidth}
        \includegraphics[width=\textwidth, trim=5 5 5 5, clip]{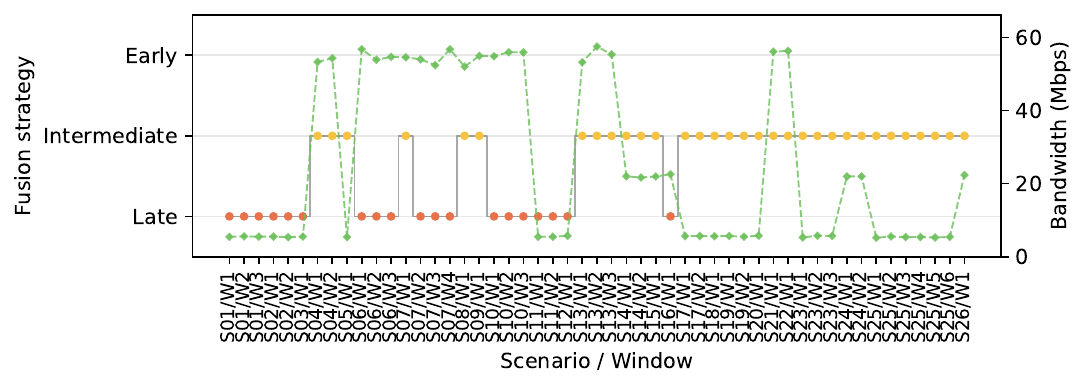}
        \caption{Llama3.3 70B.}
        \label{fig:fusion_selection_llama}
    \end{subfigure}
    \vspace{0.5em}
    \begin{subfigure}{0.48\textwidth}
        \includegraphics[width=\textwidth, trim=5 5 5 5, clip]{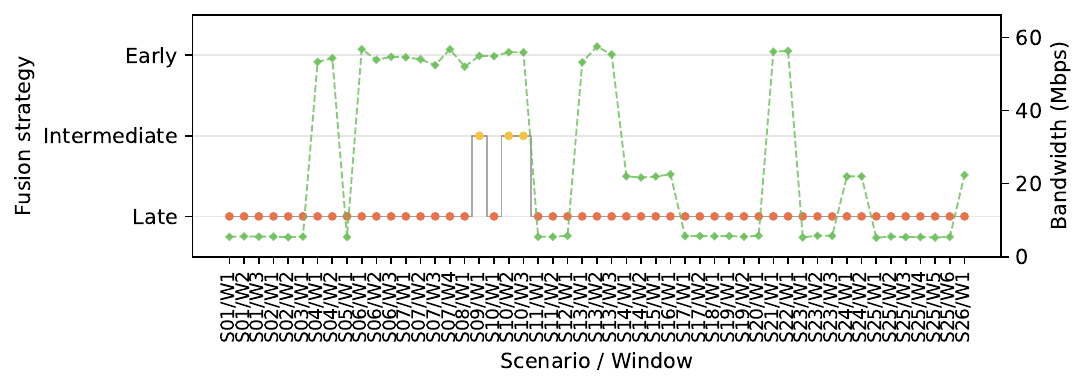}
        \caption{GPT-OSS-120B.}
        \label{fig:fusion_selection_gpt120b}
    \end{subfigure}
    \hfill
    \begin{subfigure}{0.48\textwidth}
        \includegraphics[width=\textwidth, trim=5 5 5 5, clip]{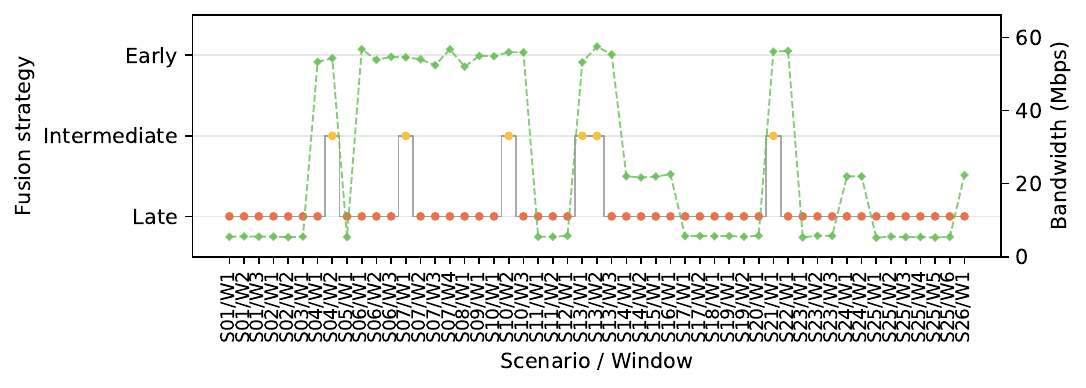}
        \caption{GPT-OSS-20B.}
        \label{fig:fusion_selection_gpt20b}
    \end{subfigure}
    \caption{\oursystem-selected fusion strategy vs. available bandwidth.}
    \label{fig:fusion_strategy}
\end{figure*}

\begin{figure}[thbp]
    \centering
    \includegraphics[width=\linewidth, trim=5 5 5 5, clip]{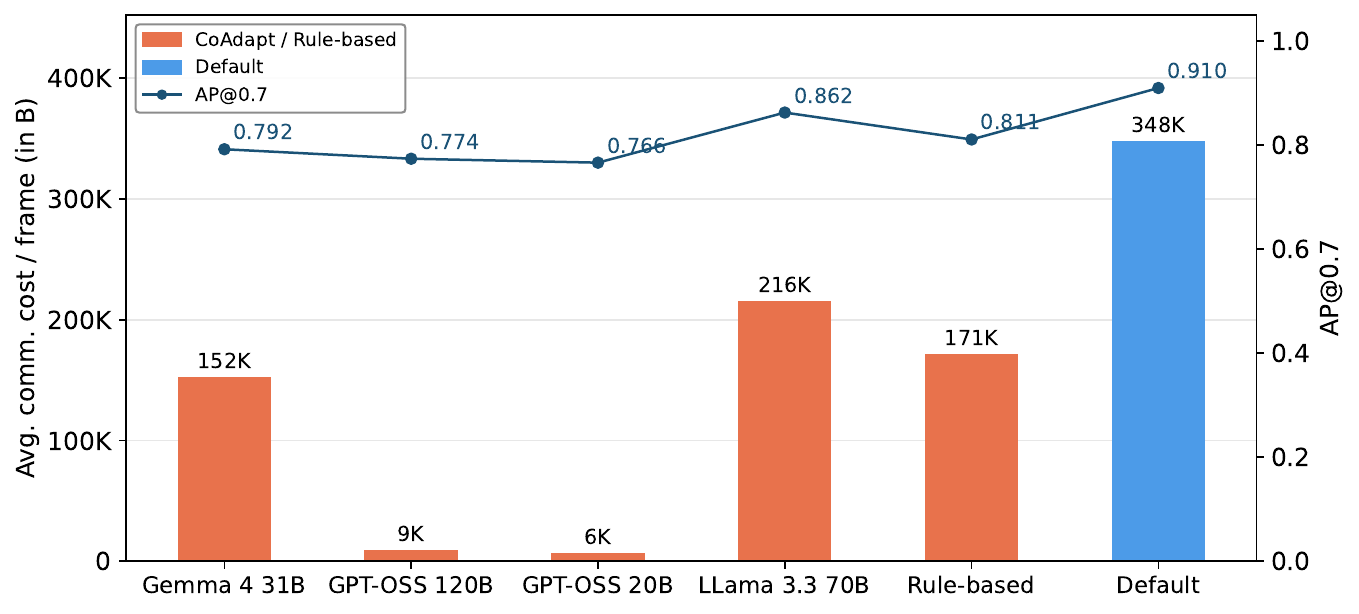}
    \caption{Average communication cost / frame vs. average precision.}
    \label{fig:approach_comparison}
\end{figure}

To illustrate that participant reduction reflects strategic spatial reasoning rather than indiscriminate exclusion, Figure~\ref{fig:cav_positions} shows the spatial configuration of selected and excluded CAVs for a representative scenario\footnote{Additional figures for all scenarios are available in \oursystem's code repository.}. The ego vehicle is positioned at the center of its 45m communication range, which corresponds to the communication range used in the OPV2V dataset.
Figure~\ref{fig:cav_positions} shows that the \oursystem-selected CAVs are often those that provide geometrically complementary coverage, while excluded CAVs tend to be spatially clustered near already-selected participants, introducing redundant observations without coverage gain.
This pattern is consistent across three models: Gemma 4, GPT-OSS 120B and Llama 3.3.
All three exclude the CAV furthest from providing unique coverage, though they differ in which specific agents they retain.
This shows that the LLMs successfully internalize spatial reasoning over the scene descriptions produced by the SAM, selecting participants based on coverage complementarity rather than proximity or availability alone.
GPT-OSS 20B, on the other hand, retains spatially redundant CAVs that provide overlapping coverage, suggesting that smaller models may lack the spatial reasoning capacity to reliably identify complementary configurations.
This points to a limitation of purely LLM-driven selection, and motivates the integration of a deterministic fallback component as part of future work to handle cases where LLM reasoning produces suboptimal participant sets.

\subsection{Precision Evaluation under Dynamic Bandwidth}

\begin{figure*}[t]
    \centering
    \begin{subfigure}{0.48\textwidth}
        \includegraphics[width=\textwidth]{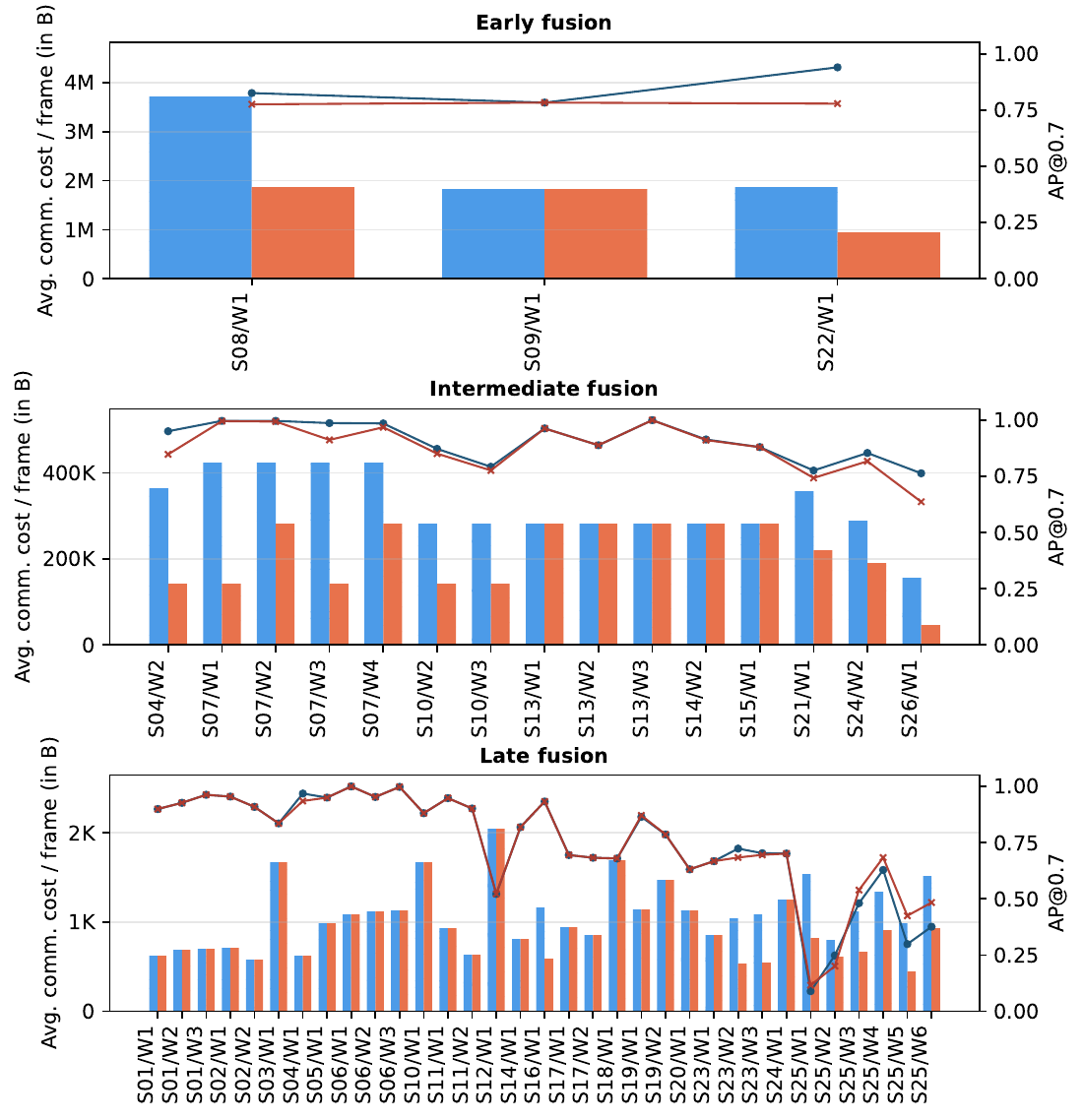}
        \caption{Gemma 4 31B.}
        \label{fig:comm_cost_gemma}
    \end{subfigure}
    \hfill
        \begin{subfigure}{0.48\textwidth}
        \includegraphics[width=\textwidth]{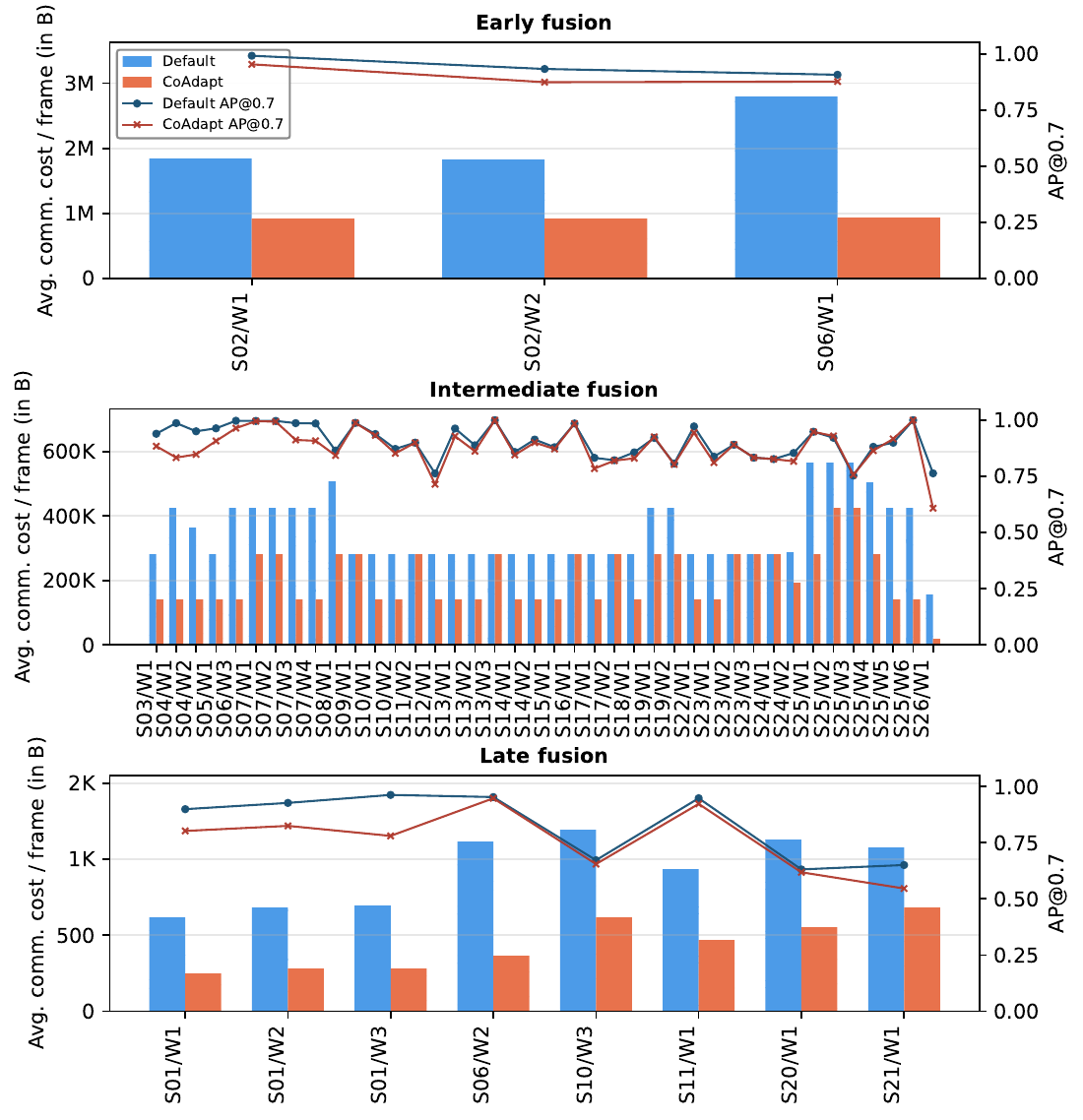}
        \caption{Llama 3.3 70B.}
        \label{fig:comm_cost_llama}
    \end{subfigure}
    \vspace{0.5em}
    \begin{subfigure}{0.48\textwidth}
        \includegraphics[width=\textwidth,]{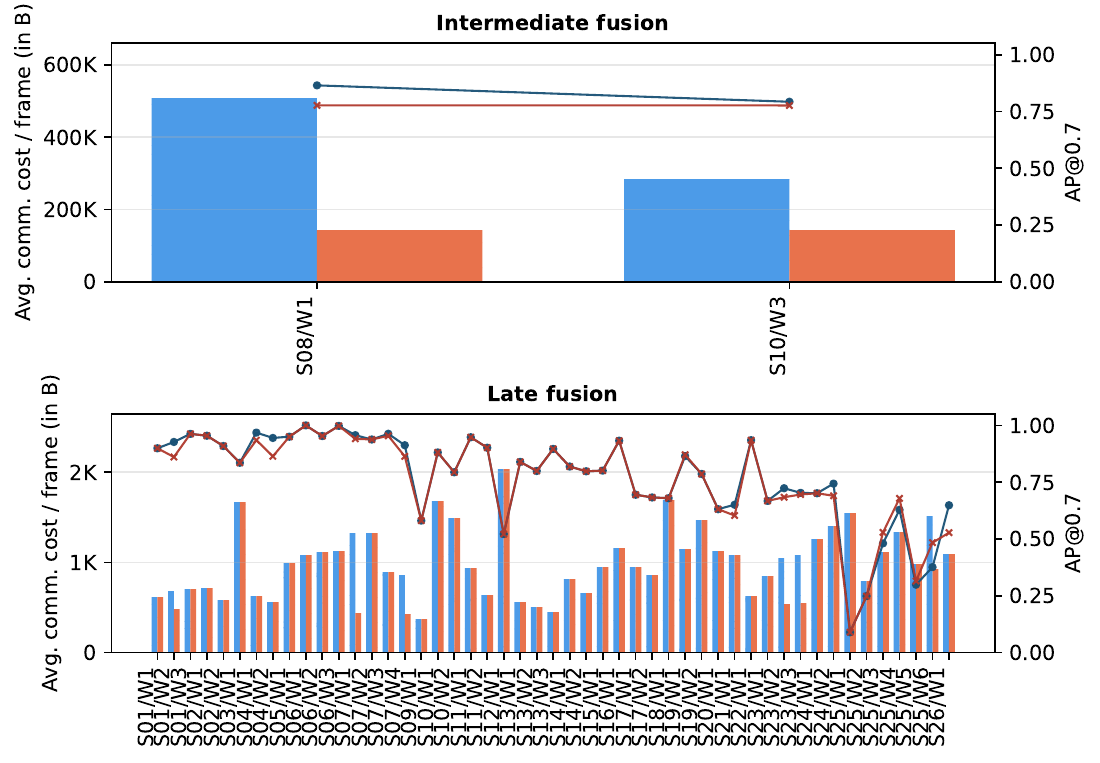}
        \caption{GPT-OSS 20B.}
        \label{fig:comm_cost_gpt20b}
    \end{subfigure}
    \hfill
    \begin{subfigure}{0.48\textwidth}
        \includegraphics[width=\textwidth]{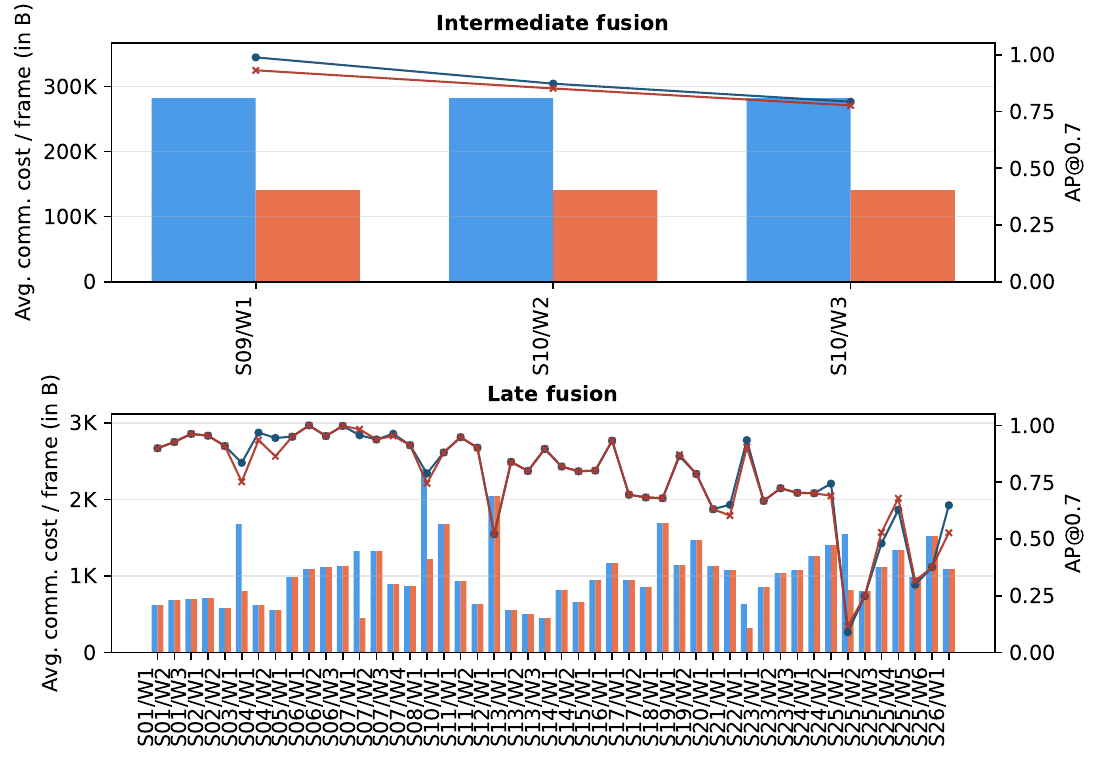}
        \caption{GPT-OSS 120B.}
        \label{fig:comm_cost_gpt120b}
    \end{subfigure}
    \caption{Average communication cost / frame / fusion strategy vs. precision.}
    \label{fig:comm_cost}
\end{figure*}

We now analyze how \oursystem adapts the fusion strategy under dynamic network conditions.
Figure~\ref{fig:fusion_strategy} shows the fusion strategy selected by each LLM across all evaluation scenarios, alongside the simulated available bandwidth. 

\oursystem dynamically switches between fusion paradigms across control cycles, with the selected algorithm reflecting the trade-off between available bandwidth and detection precision.
We can notice that Gemma 4 and Llama3.3 models exhibit the richest adaptive behavior, switching across all three fusion paradigms in response to bandwidth variations.
They often select intermediate fusion during high-bandwidth periods and fall back to late fusion when bandwidth is constrained. 
Early fusion is selected by Gemma 4 at only three control cycles, when bandwidth is high enough to accommodate sending raw point clouds over the network. 

GPT-OSS models, on the other hand, show a more conservative approach, selecting almost exclusively late fusion across scenarios.
They rarely select intermediate fusion, and never select early fusion.
This conservative behavior is consistent with the high communication cost of early fusion in practice, which makes it an unattractive choice under realistic wireless network constraints. 
This suggests a tendency to default to lower-cost, lower-risk options under uncertainty. 
Nonetheless, the strategy selections remain generally coherent with the network conditions: late fusion is a valid and communication-efficient choice, and the models do not select early fusion when bandwidth is low.

We next evaluate the communication cost-precision tradeoff.
For comparison purposes, we implement a rule-based approach that selects intermediate fusion when enough bandwidth (more than 20~Mbps) is available, and late fusion otherwise.
We compare \oursystem against the rule-based approach, and existing approaches that use a static intermediate fusion policy regardless of bandwidth fluctuations.
We measure the average precision at an Intersection over Union (IoU) threshold of 0.7 (AP@0.7), which constitutes a standard metric for 3D object detection that measures the fraction of detected objects whose predicted bounding boxes sufficiently overlap with ground truth annotations.
Figure~\ref{fig:approach_comparison} shows the average communication cost per frame vs. the average precision for \oursystem with the 4 used LLMs against the rule-based and default approach.
Because GPT-OSS models almost always choose late fusion, their average communication cost is significantly low: both models have an average cost of less than 10 KB / frame. This is expected, as late fusion models are very communication-efficient, transmitting only detection boxes in the form of textual data.
However, they achieve the lowest precisions, at an AP@0.7 of 0.766 and 0.774 for GPT-OSS 20B and GPT-OSS 120B, respectively. 
Gemma 4, on the other hand, achieves a favorable communication cost-precision tradeoff, with an AP@0.7 of 0.792 against an average communication cost of 152.2 KB/frame.
This constitutes an $11\%$ reduction in communication load at a cost of a $2.3\%$ precision drop compared to rule-based approaches, which represents a comparable precision at lower communication cost.
In terms of precision, Llama 3.3 achieves the best precision at an AP@0.7 of 0.862, for a communication cost that is $38\%$ lower than default approaches.
These results demonstrate that LLM-driven fusion policy selection offers a favorable and configurable tradeoff between communication cost and detection precision.

We further show the communication cost and AP@0.7 of \oursystem  compared to the default full-participation baseline, broken down by fusion paradigm and model.
Figure~\ref{fig:comm_cost} groups scenarios by the fusion algorithm selected by the \oursystem at each control window.
Across all models and fusion paradigms, \oursystem-selected participant subsets consistently reduce communication cost compared to the default baseline while maintaining comparable detection precision. In intermediate fusion windows, \oursystem achieves substantial cost reductions of $50\%$ with GPT-OSS models, $42\%$ with the Llama model, and $38\%$ with the Gemma model, by pruning redundant participants.
AP@0.7 values are comparable between \oursystem and default configuration, confirming that participant pruning does not meaningfully degrade precision.
A similar pattern holds in late fusion windows, where the already low communication cost of late fusion is further reduced by selecting a smaller participant subset, again without significant precision loss compared to the default approach.

Note that precision drops in scenarios S12 and S25 are attributable to scene-level difficulty inherent to those scenarios (which is demonstrated by a precision drop for the default approach), rather than suboptimal participant selection by \oursystem.
This confirms that the LLM policy does not introduce additional precision risk beyond what the underlying fusion models already exhibit under challenging conditions.

\subsection{Limitations}
While \oursystem demonstrates promising results in using LLMs in adaptive collaborative perception, several limitations must be acknowledged:
\newline
\noindent \textbf{Synthetic Bandwidth Model.} The bandwidth conditions used in this evaluation are synthetically derived from scenario timestamps in OPV2V, mapping shift hours to predefined tiers with sinusoidal fluctuations.
While this model captures the temporal structure of industrial network load, it does not reflect the full complexity of real IIoT wireless environments, which exhibit bursty interference, spatial variation, and protocol-level dynamics. Validating \oursystem against real network traces or a real testbed is part of our future work.

\noindent \textbf{Dataset Constraints.} The evaluation is limited by the scarcity of collaborative perception datasets supporting high CAV counts.
OPV2V provides up to six simultaneous agents per scenario, which, while sufficient to demonstrate the adaptive behavior of \oursystem, does not reflect the scale of large industrial swarm deployments where tens of robots may operate concurrently.
The decision space grows exponentially with the number of agents, as described in Section~\ref{sec:problem}, and scenarios with larger participant sets are the ones where the LLM's reasoning ability can be showcased against static approaches. 
The development of large-scale multi-agent perception datasets is an open challenge for the collaborative perception community~\cite{wang2025collaborative}, and future evaluation of our framework will depend on the availability of such datasets.

\noindent \textbf{Absence of a Deterministic Fallback.} Our system currently relies entirely on LLM reasoning for participant selection and fusion strategy decisions.
Even though our evaluation shows that tested LLMs produced valid outputs in response to all queries, no deterministic fallback exists for edge cases where the model produces suboptimal or invalid output.
As demonstrated by GPT-OSS 20B's failure to identify spatially redundant CAVs in certain scenarios, smaller models may lack the spatial reasoning capacity required for reliable operation.
A hybrid architecture combining LLM-driven reasoning with a rule-based safety layer would improve robustness and is a natural direction for future work.

\noindent \textbf{LLM Inference Latency.}  While \oursystem focuses on demonstrating the feasibility and capabilities of LLM-driven autonomic adaptation for collaborative perception, inference latency represents a practical constraint that must be addressed before deployment in real-time control loops.
Our evaluation showed that the mean LLM response times range from 21 seconds for Gemma 4 31B to 44 seconds for Llama 3.3 70B.
Several approaches exist to reduce LLM inference latency, including model quantization~\cite{lin2024awq}, knowledge distillation into smaller surrogate models~\cite{distillation}, and speculative decoding~\cite{xia2024unlocking}, and their integration into the control loop is left as future work.

\noindent \textbf{}


%% file: conclusion.tex
This paper presents \oursystem, an adaptive collaborative perception framework for IIoT robotic swarms.
We rely on Large Language Models as runtime fusion controllers, jointly selecting the participant subset and fusion algorithm at each control cycle based on the current spatial configuration and network state. 
We implement a Scene Abstraction Module to bridge the gap between raw LiDAR point cloud data and LLM reasoning by converting heterogeneous physical observations into structured natural language scene descriptions.
\oursystem instantiates a MAPE-K control loop over the collaborative fusion pipeline, enabling self-adaptive behavior without task-specific training or exhaustive online search.
Experimental evaluation on the OPV2V benchmark across 25 scenarios demonstrates that \oursystem consistently reduces communication cost compared to the static full-participation baseline, while maintaining detection precision comparable to the underlying fusion models. 

Our future work will focus on three extensions. 
First, we will implement the Network Control Layer, enabling active adaptation of the wireless infrastructure to support higher robot participation when application precision requirements cannot be met through fusion policy adjustment alone.
Second, we will introduce a constraint validation layer that verifies LLM outputs against physical and operational constraints before execution, coupled with a rule-based fallback policy that activates when the LLM produces invalid or low-confidence decisions. 
Third, we will evaluate the framework on physical robotic platforms and real-world network traces to assess the sim-to-real transferability of our approach.

%% file: main.bbl
\begin{thebibliography}{10}
\providecommand{\url}[1]{#1}
\csname url@samestyle\endcsname
\providecommand{\newblock}{\relax}
\providecommand{\bibinfo}[2]{#2}
\providecommand{\BIBentrySTDinterwordspacing}{\spaceskip=0pt\relax}
\providecommand{\BIBentryALTinterwordstretchfactor}{4}
\providecommand{\BIBentryALTinterwordspacing}{\spaceskip=\fontdimen2\font plus
\BIBentryALTinterwordstretchfactor\fontdimen3\font minus
  \fontdimen4\font\relax}
\providecommand{\BIBforeignlanguage}[2]{{%
\expandafter\ifx\csname l@#1\endcsname\relax
\typeout{** WARNING: IEEEtran.bst: No hyphenation pattern has been}%
\typeout{** loaded for the language `#1'. Using the pattern for}%
\typeout{** the default language instead.}%
\else
\language=\csname l@#1\endcsname
\fi
#2}}
\providecommand{\BIBdecl}{\relax}
\BIBdecl

\bibitem{vermesan2022internet}
O.~Vermesan, A.~Br{\"o}ring, E.~Tragos, M.~Serrano, D.~Bacciu, S.~Chessa,
  C.~Gallicchio, A.~Micheli, M.~Dragone, A.~Saffiotti \emph{et~al.},
  ``{Internet of Robotic Things--Converging Sensing/Actuating,
  Hyperconnectivity, Artificial Intelligence and IoT Platforms},'' in
  \emph{Cognitive hyperconnected digital transformation}.\hskip 1em plus 0.5em
  minus 0.4em\relax River Publishers, 2022, pp. 97--155.

\bibitem{debie2023swarm}
E.~Debie, K.~Kasmarik, and M.~Garratt, ``{Swarm Robotics: A Survey from a
  Multi-tasking Perspective},'' \emph{ACM Computing Surveys}, vol.~56, no.~2,
  pp. 1--38, 2023.

\bibitem{wang2020v2vnet}
T.-H. Wang, S.~Manivasagam, M.~Liang, B.~Yang, W.~Zeng, J.~Tu, and R.~Urtasun,
  ``{V2VNet: Vehicle-to-Vehicle Communication for Joint Perception and
  Prediction},'' in \emph{European Conference on Computer Vision (ECCV)}, 2020.

\bibitem{chen2019fcooper}
Q.~Chen, X.~Ma, S.~Tang, J.~Guo, Q.~Yang, and S.~Fu, ``{F-Cooper: Feature Based
  Cooperative Perception for Autonomous Vehicle Edge Computing System Using 3D
  Point Clouds},'' in \emph{ACM/IEEE Symposium on Edge Computing (SEC)}, 2019.

\bibitem{xu2022cobevt}
R.~Xu, Z.~Tu, H.~Xiang, W.~Shao, B.~Zhou, and J.~Ma, ``{CoBEVT: Cooperative
  Bird's Eye View Semantic Segmentation with Sparse Transformers},'' in
  \emph{Conference on Robot Learning (CoRL)}, 2022.

\bibitem{xu2022opv2v}
R.~Xu, H.~Xiang, X.~Xia, X.~Han, J.~Li, and J.~Ma, ``{OPV2V: An Open Benchmark
  Dataset and Fusion Pipeline for Perception with Vehicle-to-Vehicle
  Communication},'' in \emph{IEEE International Conference on Robotics and
  Automation (ICRA)}, 2022.

\bibitem{liu2020who2com}
Y.-C. Liu, J.~Tian, C.-Y. Ma, N.~Glaser, C.-W. Kuo, and Z.~Kira, ``{Who2com:
  Collaborative Perception via Learnable Handshake Communication},'' in
  \emph{2020 IEEE International Conference on Robotics and Automation
  (ICRA)}.\hskip 1em plus 0.5em minus 0.4em\relax IEEE, 2020, pp. 6876--6883.

\bibitem{liu2020when2com}
Y.-C. Liu, J.~Tian, N.~Glaser, and Z.~Kira, ``{When2com: Multi-agent Perception
  via Communication Graph Grouping},'' in \emph{Proceedings of the IEEE/CVF
  Conference on computer vision and pattern recognition}, 2020, pp. 4106--4115.

\bibitem{hu2022where2comm}
Y.~Hu, S.~Fang, Z.~Lei, Y.~Zhong, and S.~Chen, ``{Where2comm:
  Communication-Efficient Collaborative Perception via Spatial Confidence
  Maps},'' in \emph{Advances in Neural Information Processing Systems
  (NeurIPS)}, 2022.

\bibitem{yang2023how2comm}
D.~Yang, K.~Yang, Y.~Wang, J.~Liu, Z.~Xu, R.~Yin, P.~Zhai, and L.~Zhang,
  ``{How2comm: Communication-Efficient and Collaboration-pragmatic Multi-agent
  Perception},'' \emph{Advances in Neural Information Processing Systems},
  vol.~36, pp. 25\,151--25\,164, 2023.

\bibitem{golpayegani2018participant}
F.~Golpayegani, Z.~Sahaf, I.~Dusparic, and S.~Clarke, ``{Participant Selection
  for Short-term Collaboration in Open Multi-agent Systems},'' \emph{Simulation
  Modelling Practice and Theory}, vol.~83, pp. 149--161, 2018.

\bibitem{li2021learning}
Y.~Li, S.~Ren, P.~Wu, S.~Chen, C.~Feng, and W.~Zhang, ``{Learning Distilled
  Collaboration Graph for Multi-agent Perception},'' \emph{Advances in Neural
  Information Processing Systems}, vol.~34, pp. 29\,541--29\,552, 2021.

\bibitem{hadidi2018distributed}
R.~Hadidi, J.~Cao, M.~Woodward, M.~S. Ryoo, and H.~Kim, ``{Distributed
  Perception by Collaborative Robots},'' \emph{IEEE Robotics and Automation
  Letters}, vol.~3, no.~4, pp. 3709--3716, 2018.

\bibitem{han2023collaborative}
Y.~Han, H.~Zhang, H.~Li, Y.~Jin, C.~Lang, and Y.~Li, ``{Collaborative
  Perception in Autonomous Driving: Methods, Datasets, and Challenges},''
  \emph{IEEE Intelligent Transportation Systems Magazine}, vol.~15, no.~6, pp.
  131--151, 2023.

\bibitem{xu2022v2xvit}
R.~Xu, H.~Xiang, Z.~Tu, X.~Xia, M.-H. Yang, and J.~Ma, ``{V2X-ViT:
  Vehicle-to-Everything Cooperative Perception with Vision Transformer},''
  \emph{European Conference on Computer Vision (ECCV)}, 2022.

\bibitem{lei2022latency}
Z.~Lei, S.~Ren, Y.~Hu, W.~Zhang, and S.~Chen, ``{Latency-aware Collaborative
  Perception},'' in \emph{European Conference on Computer Vision}.\hskip 1em
  plus 0.5em minus 0.4em\relax Springer, 2022, pp. 316--332.

\bibitem{hu2024communication}
Y.~Hu, J.~Peng, S.~Liu, J.~Ge, S.~Liu, and S.~Chen, ``{Communication-efficient
  Collaborative Perception via Information Filling with Codebook},'' in
  \emph{Proceedings of the IEEE/CVF Conference on Computer Vision and Pattern
  Recognition}, 2024, pp. 15\,481--15\,490.

\bibitem{ding2025point}
Z.~Ding, J.~Fu, S.~Liu, H.~Li, S.~Chen, H.~Li, S.~Zhang, and X.~Zhou, ``{Point
  Cluster: A Compact Message Unit for Communication-efficient Collaborative
  Perception},'' in \emph{The Thirteenth International Conference on Learning
  Representations}, 2025.

\bibitem{xu2025cosdh}
J.~Xu, Y.~Zhang, Z.~Cai, and D.~Huang, ``{CoSDH: Communication-efficient
  Collaborative Perception via Supply-demand Awareness and Intermediate-late
  Hybridization},'' in \emph{Proceedings of the Computer Vision and Pattern
  Recognition Conference}, 2025, pp. 6834--6843.

\bibitem{alberts2024rebet}
E.~Alberts, I.~Gerostathopoulos, V.~Stoico, and P.~Lago, ``{ReBeT:
  Architecture-based Self-adaptation of Robotic Systems through Behavior
  Trees},'' in \emph{2024 IEEE International Conference on Autonomic Computing
  and Self-Organizing Systems (ACSOS)}.\hskip 1em plus 0.5em minus 0.4em\relax
  IEEE, 2024, pp. 1--10.

\bibitem{atik2024sustainability}
S.~T. Atik, D.~Grosu, and M.~Brocanelli, ``{Sustainability-aware Online Task
  and Charge Allocation for Autonomous Ground Robot Fleets},'' in \emph{2024
  IEEE International Conference on Autonomic Computing and Self-Organizing
  Systems (ACSOS)}.\hskip 1em plus 0.5em minus 0.4em\relax IEEE, 2024, pp.
  11--20.

\bibitem{schilcher2025swarmalators}
U.~Schilcher, M.~Schref, and C.~Bettstetter, ``{Swarmalators with Discrete-Time
  Coupling: Which Step Size to Choose?}'' in \emph{2025 IEEE International
  Conference on Autonomic Computing and Self-Organizing Systems (ACSOS)}.\hskip
  1em plus 0.5em minus 0.4em\relax IEEE, 2025, pp. 112--121.

\bibitem{aguzzi2025field}
G.~Aguzzi, M.~Baiardi, A.~Cortecchia, B.~Miloradovic, A.~Papadopoulos,
  D.~Pianini, and M.~Viroli, ``{A Field-based Approach for Runtime Replanning
  in Swarm Robotics Missions},'' in \emph{2025 IEEE International Conference on
  Autonomic Computing and Self-Organizing Systems (ACSOS)}.\hskip 1em plus
  0.5em minus 0.4em\relax IEEE, 2025, pp. 1--10.

\bibitem{huang2022language}
W.~Huang, P.~Abbeel, D.~Pathak, and I.~Mordatch, ``{Language Models as
  Zero-Shot Planners: Extracting Actionable Knowledge for Embodied Agents},''
  in \emph{International Conference on Machine Learning (ICML)}, 2022, pp.
  9118--9147.

\bibitem{kannan2024smart}
S.~S. Kannan, V.~L. Venkatesh, and B.-C. Min, ``S{mart-LLM: Smart Multi-agent
  Robot Task Planning using Large Language Models},'' in \emph{2024 IEEE/RSJ
  International Conference on Intelligent Robots and Systems (IROS)}.\hskip 1em
  plus 0.5em minus 0.4em\relax IEEE, 2024, pp. 12\,140--12\,147.

\bibitem{ao2025llm}
J.~Ao, F.~Wu, Y.~Wu, A.~Swiki, and S.~Haddadin, ``{LLM-as-BT-Planner:
  Leveraging LLMs for behavior tree generation in robot task planning},'' in
  \emph{2025 IEEE International Conference on Robotics and Automation
  (ICRA)}.\hskip 1em plus 0.5em minus 0.4em\relax IEEE, 2025, pp. 1233--1239.

\bibitem{liu2024vision}
S.~Liu, J.~Zhang, R.~X. Gao, X.~V. Wang, and L.~Wang, ``{Vision-language
  Model-driven Scene Understanding and Robotic Object Manipulation},'' in
  \emph{2024 IEEE 20th International Conference on Automation Science and
  Engineering (CASE)}.\hskip 1em plus 0.5em minus 0.4em\relax IEEE, 2024, pp.
  21--26.

\bibitem{yang2025autohma}
T.~Yang, P.~Feng, Q.~Guo, J.~Zhang, X.~Zhang, J.~Ning, X.~Wang, and Z.~Mao,
  ``{AutoHMA-LLM: Efficient Task Coordination and Execution in Heterogeneous
  Multi-agent Systems using Hybrid Large Language Models},'' \emph{IEEE
  Transactions on Cognitive Communications and Networking}, vol.~11, no.~2, pp.
  987--998, 2025.

\bibitem{nascimento2023selfadaptive}
N.~Nascimento, P.~Alencar, and D.~Cowan, ``{Self-adaptive Large Language Model
  (LLM)-based Multiagent Systems},'' in \emph{2023 IEEE International
  Conference on Autonomic Computing and Self-Organizing Systems Companion
  (ACSOS-C)}.\hskip 1em plus 0.5em minus 0.4em\relax IEEE, 2023, pp. 104--109.

\bibitem{chiu2025v2v}
H.-k. Chiu, R.~Hachiuma, C.-Y. Wang, S.~F. Smith, Y.-C.~F. Wang, and M.-H.
  Chen, ``{V2V-LLM: Vehicle-to-vehicle cooperative autonomous driving with
  multi-modal large language models},'' \emph{arXiv preprint arXiv:2502.09980},
  2025.

\bibitem{li2024genai}
J.~Li, M.~Zhang, N.~Li, D.~Weyns, Z.~Jin, and K.~Tei, ``{Generative AI for
  Self-Adaptive Systems: State of the Art and Research Roadmap},'' \emph{ACM
  Transactions on Autonomous and Adaptive Systems}, 2024.

\bibitem{kephart2003vision}
J.~O. Kephart and D.~M. Chess, ``{The Vision of Autonomic Computing},''
  \emph{Computer}, vol.~36, no.~1, pp. 41--50, 2003.

\bibitem{opv2v}
\url{https://mobility-lab.seas.ucla.edu/opv2v/}.

\bibitem{carla}
\url{https://carla.org/}.

\bibitem{wang2025collaborative}
N.~Wang, D.~Shang, Y.~Gong, X.~Hu, Z.~Song, L.~Yang, Y.~Huang, X.~Wang, and
  J.~Lu, ``{Collaborative Perception Datasets for Autonomous Driving: A
  Review},'' \emph{IEEE Sensors Journal}, 2025.

\bibitem{opencood}
\url{https://github.com/DerrickXuNu/OpenCOOD}.

\bibitem{chen2019cooper}
Q.~Chen, S.~Tang, Q.~Yang, and S.~Fu, ``{Cooper: Cooperative Perception for
  Connected Autonomous Vehicles based on 3D Point Clouds},'' in \emph{2019 IEEE
  39th International Conference on distributed computing systems
  (ICDCS)}.\hskip 1em plus 0.5em minus 0.4em\relax IEEE, 2019, pp. 514--524.

\bibitem{zhou2018voxelnet}
Y.~Zhou and O.~Tuzel, ``{VoxelNet: End-to-end Learning for Point Cloud based 3D
  Object Detection},'' in \emph{Proceedings of the IEEE conference on computer
  vision and pattern recognition}, 2018, pp. 4490--4499.

\bibitem{lang2019pointpillars}
A.~H. Lang, S.~Vora, H.~Caesar, L.~Zhou, J.~Yang, and O.~Beijbom,
  ``{PointPillars: Fast Encoders for Object Detection from Point Clouds},'' in
  \emph{Proceedings of the IEEE/CVF conference on computer vision and pattern
  recognition}, 2019, pp. 12\,697--12\,705.

\bibitem{zhou2018open3d}
Q.-Y. Zhou, J.~Park, and V.~Koltun, ``{Open3D: A Modern Library for 3D Data
  Processing},'' \emph{arXiv preprint arXiv:1801.09847}, 2018.

\bibitem{li2017improved}
L.~Li, F.~Yang, H.~Zhu, D.~Li, Y.~Li, and L.~Tang, ``{An Improved RANSAC for 3D
  Point Cloud Plane Segmentation Based on Normal Distribution Transformation
  Cells},'' \emph{Remote Sensing}, vol.~9, no.~5, p. 433, 2017.

\bibitem{deng2020dbscan}
D.~Deng, ``{DBSCAN Clustering Algorithm Based on Density},'' in \emph{2020 7th
  international forum on electrical engineering and automation (IFEEA)}.\hskip
  1em plus 0.5em minus 0.4em\relax IEEE, 2020, pp. 949--953.

\bibitem{lin2024awq}
J.~Lin, J.~Tang, H.~Tang, S.~Yang, W.-M. Chen, W.-C. Wang, G.~Xiao, X.~Dang,
  C.~Gan, and S.~Han, ``{AWQ: Activation-aware Weight Quantization for
  On-Device LLM Compression and Acceleration},'' \emph{Proceedings of machine
  learning and systems}, vol.~6, pp. 87--100, 2024.

\bibitem{distillation}
X.~Xu, M.~Li, C.~Tao, T.~Shen, R.~Cheng, J.~Li, C.~Xu, D.~Tao, and T.~Zhou,
  ``{A Survey on Knowledge Distillation of Large Language Models},''
  \emph{arXiv preprint arXiv:2402.13116}, 2024.

\bibitem{xia2024unlocking}
H.~Xia, Z.~Yang, Q.~Dong, P.~Wang, Y.~Li, T.~Ge, T.~Liu, W.~Li, and Z.~Sui,
  ``{Unlocking Efficiency in Large Language Model Inference: A Comprehensive
  Survey of Speculative Decoding},'' \emph{Findings of the Association for
  Computational Linguistics: ACL 2024}, pp. 7655--7671, 2024.

\end{thebibliography}
